\documentclass{article}

\usepackage[eandd, final]{neurips_2026}

\usepackage[utf8]{inputenc} 
\usepackage[T1]{fontenc}    
\usepackage{hyperref}       
\usepackage{url}            
\usepackage{booktabs}       
\usepackage{amsfonts}       
\usepackage{nicefrac}       
\usepackage{microtype}      
\usepackage{xcolor}         

\usepackage{amsmath}
\usepackage{tabularx}
\usepackage{multirow}
\usepackage{graphicx}
\usepackage{subcaption}
\usepackage{wrapfig}
\usepackage{makecell}
\usepackage{adjustbox}
\usepackage{comment}

\usepackage{tikz}
\usetikzlibrary{spy}
\usetikzlibrary{shadows}
\definecolor{cracks_magenta}{HTML}{A23B72}
\definecolor{algae_blue}{HTML}{2E86AB}
\definecolor{netcrack_orange}{HTML}{F18F01}
\definecolor{precipitation_red}{HTML}{C73E1D}
\definecolor{rust_green}{HTML}{6A994E}
\definecolor{spalling_blue}{HTML}{577A8B}
\tikzset{
    halo spy/.style={
        every spy on node/.append style={draw=white, double=#1, double distance=1pt, line width=1pt},
        every spy in node/.append style={draw=white, double=#1, double distance=1pt, line width=1pt},
        spy connection path={\draw[white, double=#1, double distance=1pt, line width=1pt] (tikzspyonnode) -- (tikzspyinnode);}
    }
}

\title{Cracks in the Foundation: A Civil Infrastructure Dataset to Challenge Vision Foundation Models}

\author{%
  Nicola Farronato\thanks{Equal Contribution} \\
  IBM Research\\
  ETH Zürich\\
  \texttt{nicola.farronato@ibm.com} \\
  \And
  Niccolo Avogaro$^*$\\
  IBM Research\\
  ETH Zürich\\
  \texttt{niccolo.avogaro1@ibm.com} \\
  \AND
  Thomas Frick \\
  IBM Research\\
  \And
  Mattia Rigotti\\
  IBM Research\\
  \And
  Rizwan Ullah Khan \\
  University of Twente\\
  \And
  Michele Magno \\
  ETH Zürich\\
  \AND
  Konrad Schindler \\
  ETH Zürich\\
  \And
  Cristiano Malossi\\
  IBM Research\\
  \And
  Florian Scheidegger \\
  IBM Research\\
}

\begin{document}

\maketitle

\begin{abstract}
  Automated structural health monitoring is essential to prevent catastrophic infrastructure failures. Precise, pixel-level defect segmentation is needed to accurately assess structural integrity, but progress in defect segmentation for civil infrastructures has been held back by an extreme scarcity of data, which requires costly expert annotation. The need for data is accentuated by algorithmic hurdles intrinsic to the problem, including center-bias and the need to rely more on shape when inspecting nearly textureless building materials. To remove the bottleneck, we introduce \textbf{Cracks in the Foundation (CiF)}, the largest and most detailed civil infrastructure (instance) segmentation dataset to date, comprising $\approx$150{,}000 high-resolution tiles meticulously curated over five years in collaboration with civil engineering experts. With the help of this unprecedented data source, we expose a blind spot of current visual AI: despite the advent of promptable Foundation Models (FMs) and Vision Language Models (VLMs), and despite the impressive abilities of today's specialised segmentation models, it turns out that dense image understanding in the built environment is nowhere near solved. 
Our evaluations indicate that even the most recent zero-shot FMs face significant challenges when deployed on real-world infrastructure and even the performance of specialised models with domain-specific supervision plateaus at  $\approx$25\% mAP. CiF establishes inspection of civil infrastructure, an elementary and seemingly easy perceptual task, as an open challenge that reveals fundamental weaknesses of present-day models trained predominantly on internet images, literally and figuratively highlighting \textit{cracks in the current foundation model paradigm}.
Dataset available at: \url{https://huggingface.co/datasets/ibm-research/cif-dataset}

\end{abstract}

\section{Introduction}
\label{sec:intro}

Automated structural health monitoring stands to revolutionize the maintenance of global civil infrastructure \citep{a_machine_learning_perspective, SHM}. In this domain, undetected defects can lead to catastrophic failures that cause massive economic damages and seriously harm humans. To assess the integrity of a structure, civil engineers require precise quantification of damage type, area, shape, and distribution. 
Technically, mapping (visible) structural damages amounts to a pixel-level segmentation problem. Given the high stakes and the complex physics of materials and their failure, high accuracy is required before a system can be deployed in the real world.

Training as well as evaluation of segmentation methods for civil infrastructure have been hobbled by data scarcity, and complicated by algorithmic peculiarities compared to general-purpose image segmentation \citep{arafin2024deep}. Datasets at the scale needed for contemporary machine learning are lacking, as crowdsourcing is not practical when annotators must have the expertise to distinguish microscopic deterioration from harmless shadows. Yet, it is not just a data problem: we will show that even with abundant data, the domain exposes fundamental weaknesses in modern visual AI architectures: the monochromatic, flat appearance of concrete means that understanding must rely on shape rather than texture, a notorious weakness of deep learning models \cite{geirhos2022imagenettrainedcnnsbiasedtexture}. The issue is further exacerbated by thin, highly non-convex target structures incompatible with the models' known center-bias \citep{zheng2024zoneevaluationrevealingspatial}.

\begin{figure}[htbp]
    \centering
    \begin{tikzpicture}[spy using outlines={rectangle, size=12mm}]
        \node[inner sep=0pt] (main_image) {\includegraphics[ width=\textwidth, keepaspectratio]{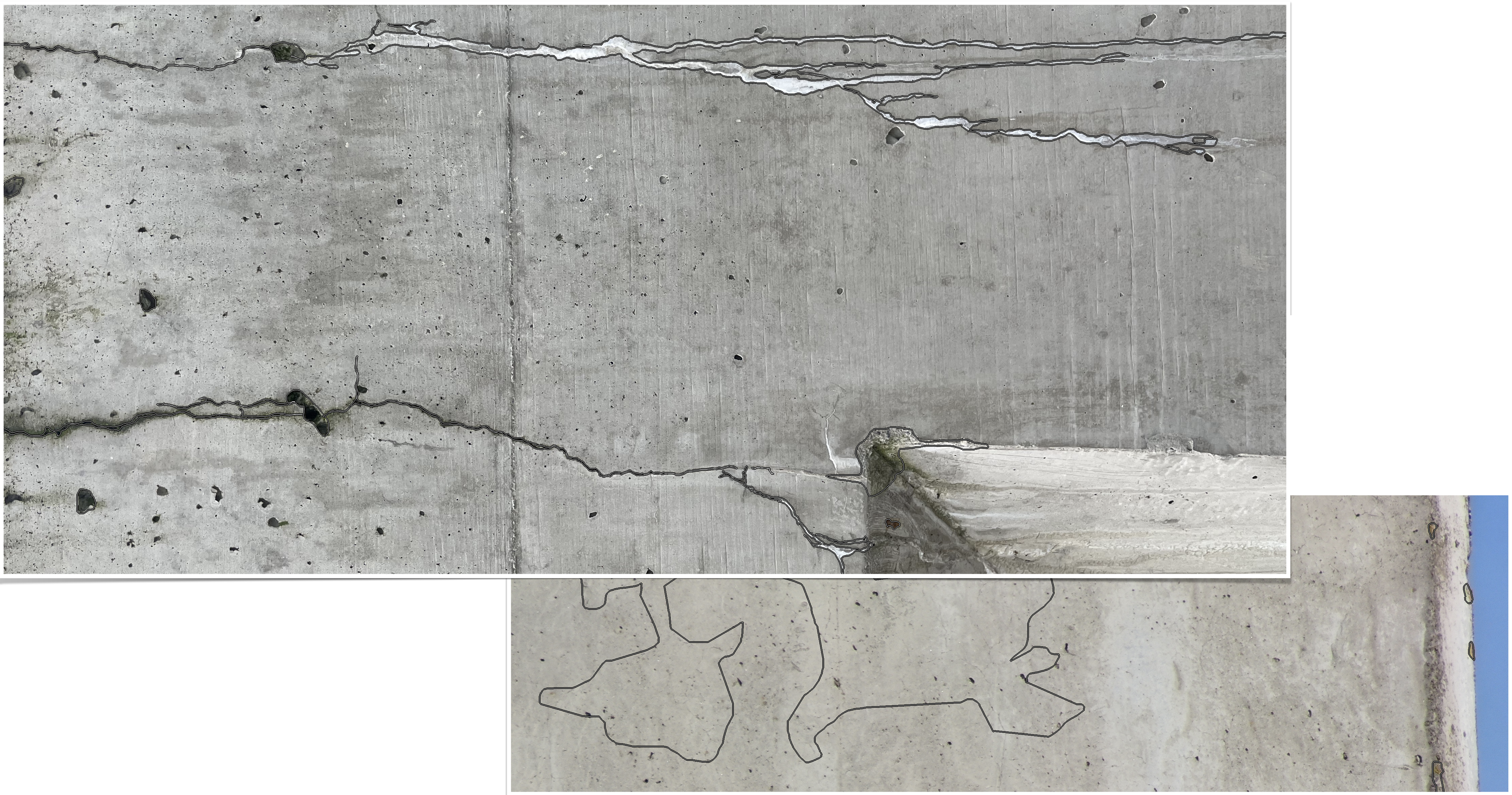}};
        
        \spy [
                magnification=4.5, size=1.5cm,
                every spy on node/.append style={fill=cracks_magenta, fill opacity=0.4} 
              ] on (-5.5,3.05) 
              in node [
                draw=cracks_magenta, line width=2pt, preaction={draw=black, line width=3pt},
              ] at (-6.2,-2.45);
        \spy [
                magnification=3, size=1.5cm,
                every spy on node/.append style={fill=precipitation_red, fill opacity=0.4} 
              ] on (-3.5,3.3) 
              in node [
                draw=precipitation_red, line width=2pt, preaction={draw=black, line width=3pt},
              ] at (-4.6,-2.45);

        \spy [
                magnification=2, size=1.5cm,
                every spy on node/.append style={fill=spalling_blue, fill opacity=0.4} 
              ] on (1.2,-0.5) 
              in node [
                draw=spalling_blue, line width=2pt, preaction={draw=black, line width=3pt},
              ] at (-3.0,-2.45);
        
        \spy [
                magnification=2, size=1.5cm,
                every spy on node/.append style={fill=netcrack_orange, fill opacity=0.4} 
              ] on (1.9,-2.3) 
              in node [
                draw=netcrack_orange, line width=2pt, preaction={draw=black, line width=3pt},
              ] at (5.75,2.83);

        \spy [
                magnification=4, size=1.5cm,
                every spy on node/.append style={fill=algae_blue, fill opacity=0.4} 
              ] on (6.5,-1.8) 
              in node [
                draw=algae_blue, line width=2pt, preaction={draw=black, line width=3pt},
              ] at (5.75,1.25);
        \spy [
                magnification=5, size=1.5cm,
                every spy on node/.append style={fill=rust_green, fill opacity=0.4} 
              ] on (1.2,-1.2) 
              in node [
                draw=rust_green, line width=2pt, preaction={draw=black, line width=3pt}, 
              ] at (5.75,-0.35);

        \node at (-6.2,-3.45) {\textcolor{cracks_magenta}{Crack}};
        \node at (-4.6,-3.45) {\textcolor{precipitation_red}{Precipitation}};
        \node at (-3.0,-3.45) {\textcolor{spalling_blue}{Spalling}};

        \node[rotate=90] at (6.8,2.83) {\textcolor{netcrack_orange}{Net-Crack}};
        \node[rotate=90] at (6.8,1.25) {\textcolor{algae_blue}{Algae}};
        \node[rotate=90] at (6.8,-0.35) {\textcolor{rust_green}{Rust}};

    \end{tikzpicture}
    \caption{Representative high-definition images from Cracks in the Foundation dataset. Due to the high resolution of the imagery and the fine scale of the structural anomalies, magnified callouts are utilized to visualize specific defect classes: Crack (magenta), Precipitation (red), Spalling (blue), Net-Crack (orange), Algae (cyan), and Rust (green).}
    \label{fig:zoomed_defects}
\end{figure}
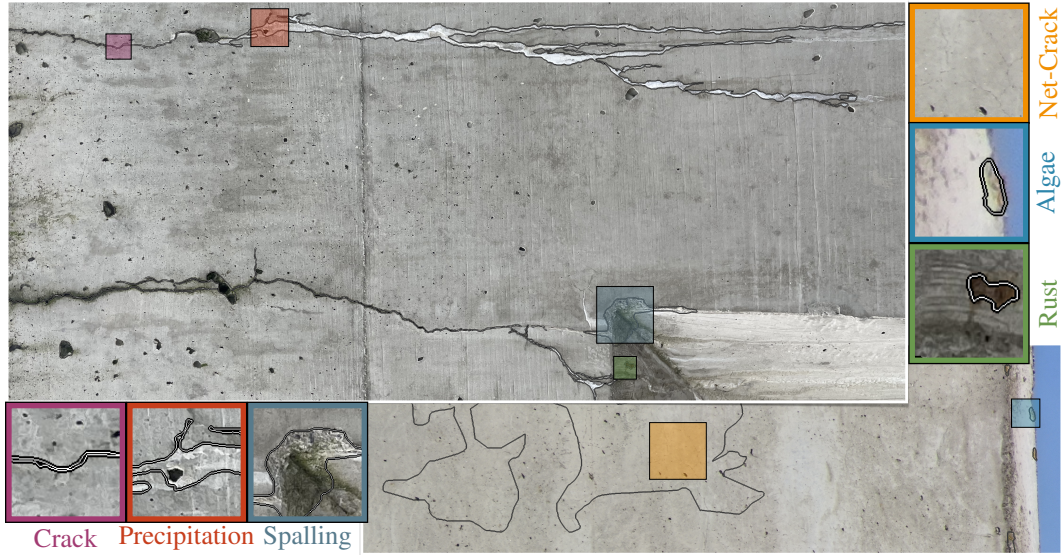

To plug this critical gap, we introduce Cracks in the Foundation (CiF), illustrated in Figure~\ref{fig:zoomed_defects}, a civil infrastructure dataset with instance annotations that is the largest of its kind in terms of both volume and resolution, consisting of $\approx$150{,}000 tiles and $\approx$250{,}000 fine-grained masks. Curated and meticulously annotated over five years of direct collaboration between AI industry partners and civil engineering experts, CiF captures the full scale range and visual complexity of infrastructure ``in the wild.'' By open-sourcing this unprecedented collection, we provide a rigorous, analysis-ready benchmark that exposes the limits of modern vision models. We hope to challenge the community and to ultimately foster the development of visual AI capable of safeguarding our physical environment.

Cracks in the Foundation especially point at an issue that has been largely overlooked in the recent AI boom. The advent of Vision Language Models \cite{alayrac2022flamingovisuallanguagemodel, liu2023visualinstructiontuning, bai2023qwenvlversatilevisionlanguagemodel, li2024llavanextinterleavetacklingmultiimagevideo, mckinzie2024mm1methodsanalysis} (VLMs), especially those capable of visual grounding \cite{peng2023kosmos2groundingmultimodallarge, deitke2024molmopixmoopenweights, you2023ferretrefergroundgranularity, rasheed2024glammpixelgroundinglarge, lai2024lisareasoningsegmentationlarge, bai2025qwen3vltechnicalreport, clark2026molmo2openweightsdata}, and of promptable Foundation Models \cite{kirillov2023segment, ravi2024sam2segmentimages, carion2026sam3segmentconcepts, liu2024groundingdinomarryingdino} (FMs) has to some degree created the illusion that dense image understanding is solved \cite{avogaro2025telleffectivelypromptingvisionlanguage}. Applying these models to the civil infrastructure domain reveals a serious, unexpectedly strong limitation: zero-shot generalization falls remarkably short. Cutting-edge FMs, trained predominantly on internet-scraped image collections like COCO \cite{lin2014microsoft} or LAION \citep{schuhmann2022laion5bopenlargescaledataset}, seem to lack understanding of the features and scale variations needed in civil engineering applications. Experiments with our benchmark demonstrate that state-of-the-art promptable models fail to identify structural anomalies, and even domain-specific, supervised models plateau at $\approx$25\% mAP. In other words, segmenting civil infrastructure degradations is, despite its straightforward nature, an open problem for present-day visual AI and challenges current solutions, literally revealing \textit{cracks in the foundation} model paradigm.

\section{Related Works}
\label{sec:ralated}
Large-scale benchmarks have catalyzed modern computer vision \citep{russakovsky2015imagenet, schuhmann2022laion5bopenlargescaledataset}, with dense prediction datasets like MSCOCO \cite{lin2014microsoft}, ADE20K \cite{ADE20K}, and SA-1B \cite{kirillov2023segment} demonstrating that high-quality instance masks are essential for training robust object detection and segmentation architectures. However, civil infrastructure inspection remains a highly specialized outlier. In this domain, generic features consistently fail to capture critical, fine-grained anomalies—such as subtle crack patterns or complex concrete deterioration—under variable real-world conditions.

To address these domain-specific needs, the civil infrastructure landscape has seen a proliferation of datasets, as comprehensively surveyed by Bianchi and Hebdon \cite{bianchi2022visual}, who cataloged 86 datasets up to 2022. These resources have evolved through levels of increasing annotation complexity, although the vast majority remain restricted to classification or semantic segmentation tasks. Early efforts primarily provided image-level or patch-level labels; for example, CDS \cite{CDS} and SDNET2018 \cite{SDNET2018} focused on binary crack detection in concrete patches. More comprehensive multi-class datasets like MCDS \cite{MCDS} and multi-label benchmarks like CODEBRIM \cite{CODEBRIM} expanded these to multiple reinforced concrete defect types. Pixel-level semantic segmentation has further matured this landscape, with single-class crack datasets like UAV75 \cite{UAV75} and CrackSeg9k \cite{CrackSeg9k} providing large-scale binary masks. Recent state-of-the-art resources such as S2DS \cite{S2DS} and dacl10k \cite{dacl10k} represent the current pinnacle of this progression, with dacl10k offering 9,920 images across 19 classes for semantic bridge damage segmentation.

Despite this growth, instance segmentation remains significantly underdeveloped in the civil infrastructure landscape. While semantic segmentation identifies defect types, it does not distinguish between separate occurrences, which is critical for damage quantification and structural health tracking. The few existing instance-level datasets suffer from substantial limitations in terms of scale, resolution, and class diversity. For instance, MCrack1300~\cite{ye2024sam} provides instance masks for only 1,300 masonry images, while CConCrack~\cite{ataei2025data} offers a limited core set of 400 images for cracks and spalls. 
Most existing instance datasets focus on a single defect class (predominantly cracks) and lack the large-scale volume of images necessary to train contemporary instance segmentation models effectively.

Table \ref{tab:ds_comparison} provides a comparative summary of the most influential datasets in this field alongside our proposed work, reporting both native full-scene parameters and derived patch-level metrics. Among classification benchmarks, SDNET2018 \cite{SDNET2018} and CODEBRIM \cite{CODEBRIM} offer high native resolutions (up to 24 MPx) but rely on low-resolution patches ($\le 0.06$ MPx) for primary tasks. In semantic segmentation, CrackSeg9k \cite{CrackSeg9k} aggregates over 2,000 images from ten sources into a 0.16 MPx patch benchmark, while dacl10k \cite{dacl10k} remains the largest full-scene resource with 9,920 images at 3.10 MPx. Current instance-level efforts like CConCrack \cite{ataei2025data} (augmented from 400 core images) represent early steps toward defect isolation but are often limited in scale or class diversity. Our CiF dataset addresses these gaps by providing 12{,}896 high-resolution (16.66 MPx) images with 148{,}642 high-fidelity tiles (1.05 MPx), representing the first large-scale, multi-class instance segmentation benchmark for bridge inspection that preserves fine-grained detail across both release variants.

\begin{table*}[htb]
\caption{Comparison of our dataset parameters with other civil infrastructure defect datasets. For image counts and resolution, we report both the original full-scene data and the derived patches/augmented samples where applicable.}
\label{tab:ds_comparison}
\centering
\setlength{\tabcolsep}{2.2mm}
\begin{tabular}{llc cc cc}
\toprule
\multirow{2}{*}{Dataset} & \multirow{2}{*}{Task} & \multirow{2}{*}{Classes} & \multicolumn{2}{c}{Images} & \multicolumn{2}{c}{Resolution (MPx)} \\
\cmidrule(lr){4-5} \cmidrule(lr){6-7}
& & & Full & Patches & Full & Patch \\
\midrule
SDNET~\cite{SDNET2018}       & Classification        & 1  & 230    & 56{,}092 & 14.06 & 0.06 \\
CODEBRIM~\cite{CODEBRIM}     & Classification        & 5  & 1{,}590 & 7{,}733  & 24.00 & 0.05 \\
\midrule
CrackSeg9k~\cite{CrackSeg9k} & Semantic seg.         & 1  & 2{,}003  & 9{,}255  & 0.25  & 0.16 \\
S2DS~\cite{S2DS}             & Semantic seg.         & 6  & 743     & --      & 1.05  & --  \\
dacl10k~\cite{dacl10k}       & Semantic seg.         & 19 & 9{,}920  & --      & 3.10  & --  \\
\midrule
Özgenel~\cite{Ozgenel2019}   & Instance seg.         & 1  & 458     & --     & 12.19 & --  \\
MCrack1300~\cite{ye2024sam}  & Instance seg.         & 5  & 1{,}300  & --     & 0.41  & --  \\
CConCrack~\cite{ataei2025data} & Instance seg.       & 2  & 400     & 10{,}995 & 12.19 & 1.05 \\
\midrule
\textbf{CiF (Ours)} & \textbf{Instance seg.} & \textbf{6} & \textbf{12{,}896} & \textbf{148{,}642} & \textbf{16.66} & \textbf{1.05} \\
\bottomrule
\end{tabular}
\end{table*}

\section{Dataset Details}
\label{sec:dataset_details}
\subsection{Data acquisition}
\label{sec:data_acquisition}
Collected over five years in collaboration with infrastructure owners (Sund \& B\ae{}lt, Trafikverket, Zurich Kanton), the dataset focuses on European civil infrastructure, particularly bridges. Its hybrid design combines heterogeneous “in-the-wild” field data to capture real-world variability with controlled inspection setups for structured, repeatable imaging. This dual approach ensures broad structural and environmental diversity, ultimately supporting robust model training and reliable evaluation.

\paragraph{In-the-Wild Data.}
The in-the-wild subset (12.7\% of the data) was collected collaboratively by engineers and researchers during routine inspections of operational infrastructure. Images were acquired in unconstrained field conditions using a variety of consumer-grade and industrial camera systems. Consequently, this subset exhibits substantial variability in resolution, viewpoint, illumination, weather conditions, and scene composition.
This variability reflects the diversity of real-world inspection scenarios and is intentionally retained to improve model generalisation. In particular, the dataset includes a wide range of defect manifestations across different structural contexts, enabling evaluation under realistic deployment conditions.

\paragraph{Controlled Inspection Setups.}
Complementing our in-the-wild data, we utilized drone-assisted controlled acquisitions to safely gather scalable, repeatable imagery in difficult environments. To simultaneously maximize structural surface coverage and ensure sufficient pixel density for fine-grained defect detection, we employed two distinct flight modes: a scan setup and a high-resolution setup. Detailed drone and camera specifications are provided in Table~\ref{tab:drone_cams}.

\begin{table*}[htb]
  \caption{Overview of camera systems and lens configurations used in the controlled inspections.}
  \label{tab:drone_cams}
  \centering
  \small
  \setlength{\tabcolsep}{4pt}
  \begin{tabular}{@{}lllll@{}}
    \toprule
    Camera & System type & Focal length [mm] & Zoom / lens & Resolution \\
    \midrule

    DJI ZH20/ZH20T
    & Optical zoom  
    & 6.83--119.94 (opt.)  
    & 23$\times$ opt., 200$\times$ digital  
    & 5184 $\times$ 3888 \\

    DJI Zenmuse X7  
    & Prime lens (S35)  
    & 50 (used; 16/24/35/50 avail.)  
    & Fixed lens  
    & 6016 $\times$ 3376 \\

    DJI Zenmuse P1  
    & Prime lens (FF)  
    & 50 (used; 24/35/50 avail.)  
    & Fixed lens  
    & 8192 $\times$ 5460 \\

    \bottomrule
  \end{tabular}

\end{table*}

\paragraph{Scan Setup.}
Prioritizing efficient, large-scale coverage over maximal detail, the scan setup utilizes a drone flying parallel to the target surface with a perpendicular camera. Operating in downward, vertical, or upward-facing orientations, the drone typically executes a zigzag trajectory to ensure systematic coverage, simultaneously recording high-precision GPS/RTK data to enable the accurate spatial registration of the captured imagery.

\paragraph{High-Resolution Setup.}
In the high-resolution setup, the drone remains in a stationary hover position while using a gimbal-mounted camera to capture a structured grid of zoomed-in images of the target surface. This approach prioritises fine-grained visual detail. To enable reconstruction of larger surface regions, we assume no oblique views so that homography transformations are used to map local views into a shared reference frame, enabling mosaic reconstruction of the observed area. This setup is particularly suitable for detailed inspection of localized structural regions where high spatial resolution is required for defect inspection.
A representative hardware configuration includes the DJI Matrice 300 RTK\footnote{\url{https://enterprise.dji.com/matrice-300}} equipped with a Zenmuse H20T\footnote{\url{https://enterprise.dji.com/zenmuse-h20-series/specs}}, although the proposed acquisition methodology is not restricted to specific hardware and can be implemented with comparable sensor systems.

\subsection{Images, Classes and Dataset Splits}
\label{sec:images_classes_splits}

The dataset contains 12{,}896 native high-resolution images (0.23 to 42.18\,MPx, mean 16.66\,MPx) annotated for six defect categories defined jointly with civil engineers. Unlike most defect benchmarks that release pre-cropped or downsampled patches, we preserve the native resolution: hairline cracks down to a few pixels in width are polygonally annotated within multi-megapixel inspection images, alongside spatially extensive defects such as algae colonisations covering large portions of a single frame. We release two coordinated variants: \emph{Full} preserves the native resolution, while \emph{Tiled} partitions every image into $1024\times1024$ patches (148{,}642 in total) using a sliding window with a stride of 768. During tiling, we discard fragmented slivers by requiring that any cropped defect instance retains at least 5\% of its total original area, resulting in a uniform-resolution benchmark suitable for current vision backbones.

\begin{figure}[htbp]
    \centering
    \setlength{\tabcolsep}{0pt}
    \renewcommand{\arraystretch}{0}
    \newcommand{\vcenteritem}[1]{\raisebox{-0.5\height}{#1}}

    \begin{tikzpicture}[
        spy using outlines={circle, connect spies},
        halo spy/.style={
            every spy on node/.append style={draw=white, double=#1, double distance=0.3pt, line width=0.3pt},
            every spy in node/.append style={draw=white, double=#1, double distance=0.3pt, line width=0.3pt},
            spy connection path={\draw[white, double=#1, double distance=0.3pt, line width=0.3pt] (tikzspyonnode) -- (tikzspyinnode);}
        }
    ]

        \node[inner sep=0pt] (mosaic) {
            \resizebox{\textwidth}{!}{%
            \begin{tabular}{c @{\hspace{1mm}} c@{}c@{}c @{\hspace{1mm}} c@{}c@{}c @{\hspace{1mm}} c}
                
                \vcenteritem{\raisebox{2mm}{\rotatebox[origin=c]{90}{Algae}}} &
                \vcenteritem{\includegraphics[width=0.14\textwidth]{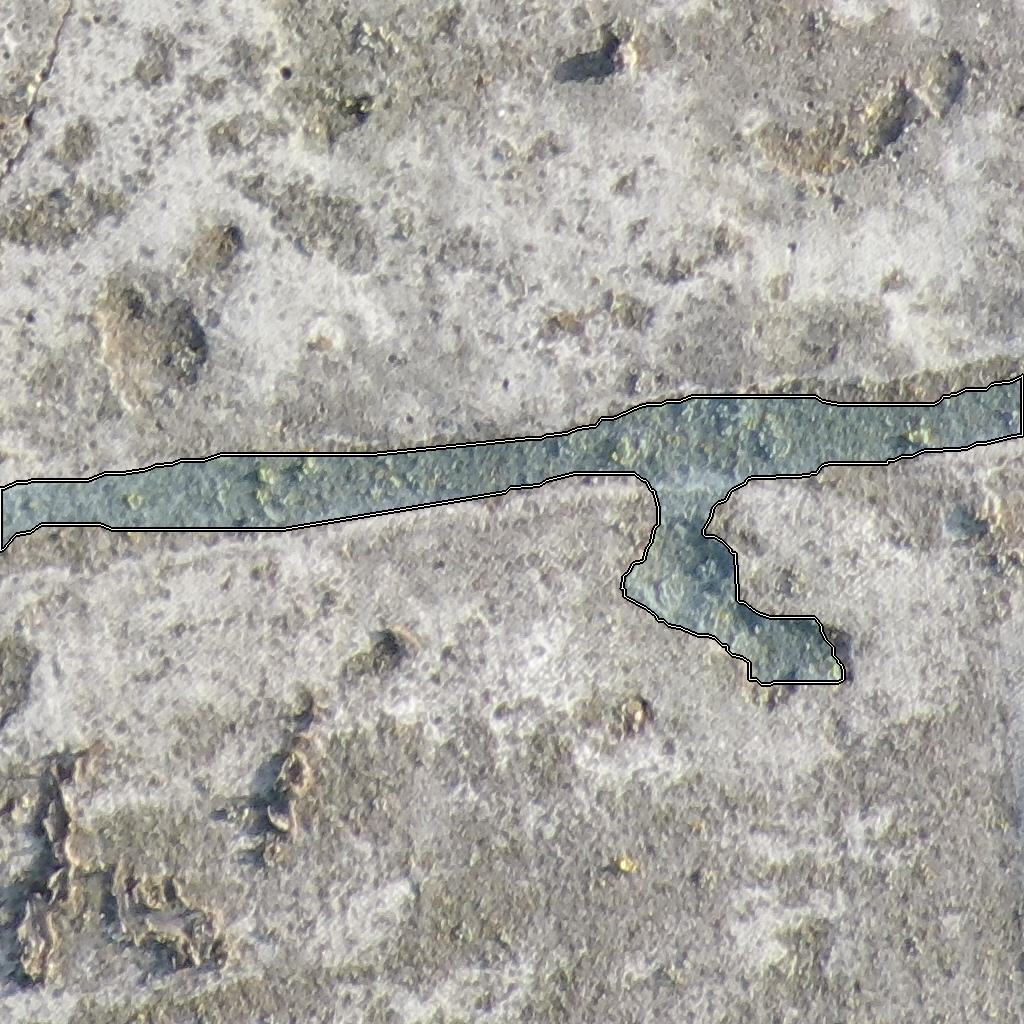}} &
                \vcenteritem{\includegraphics[width=0.14\textwidth]{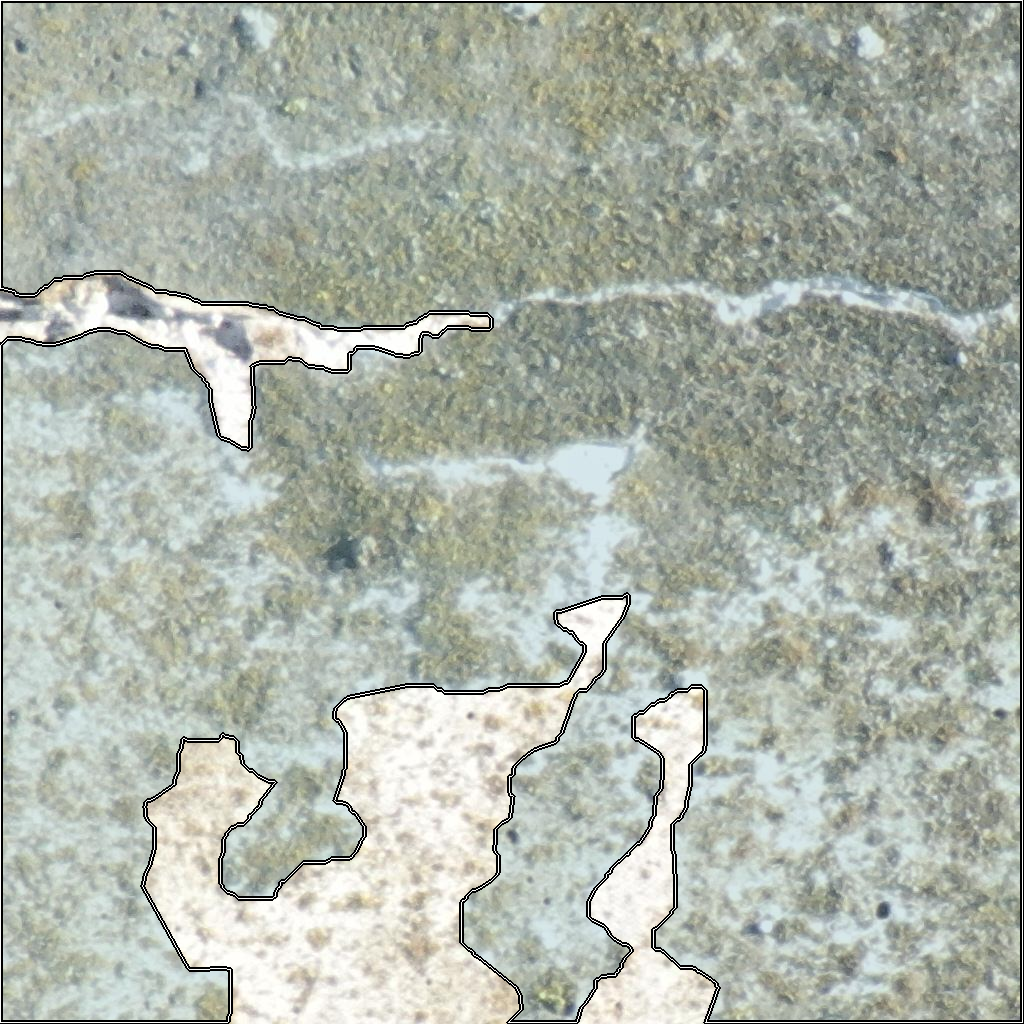}} &
                \vcenteritem{\includegraphics[width=0.14\textwidth]{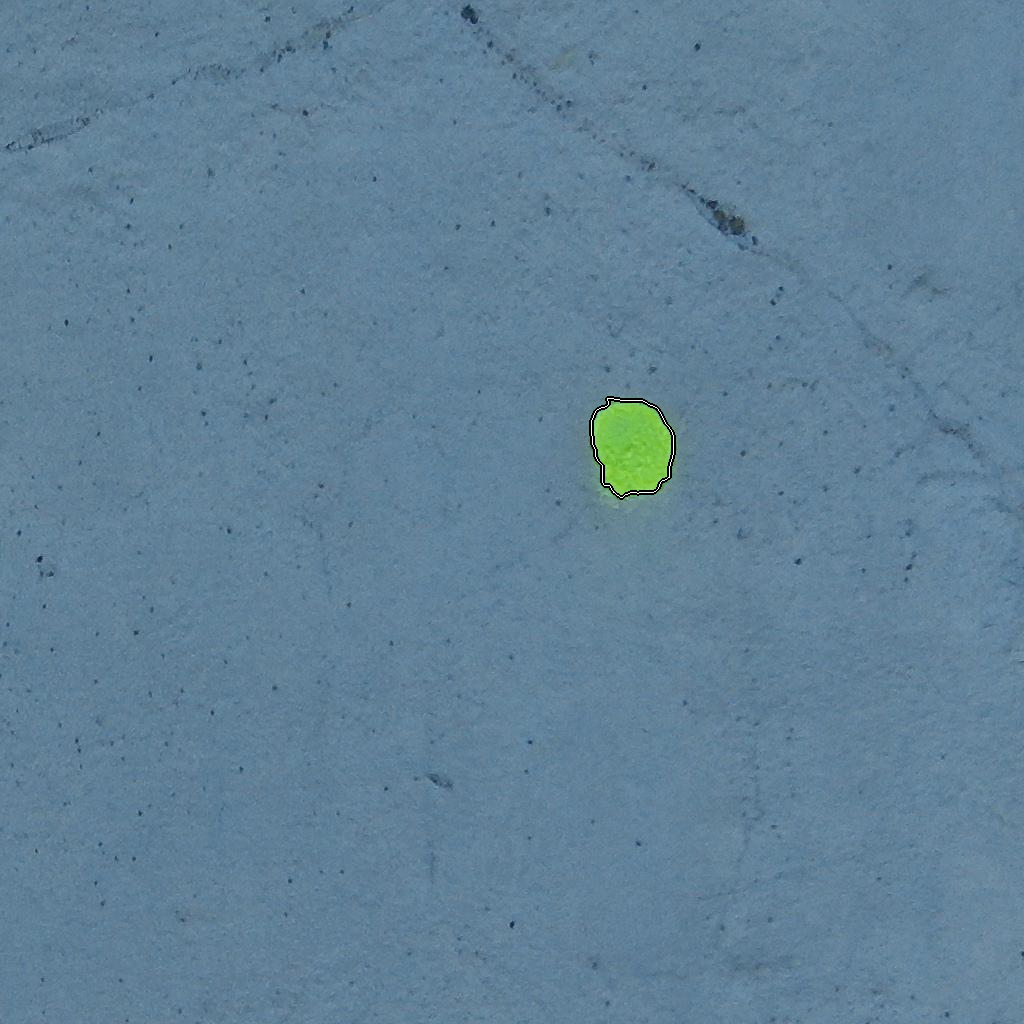}} &
                \vcenteritem{\includegraphics[width=0.14\textwidth]{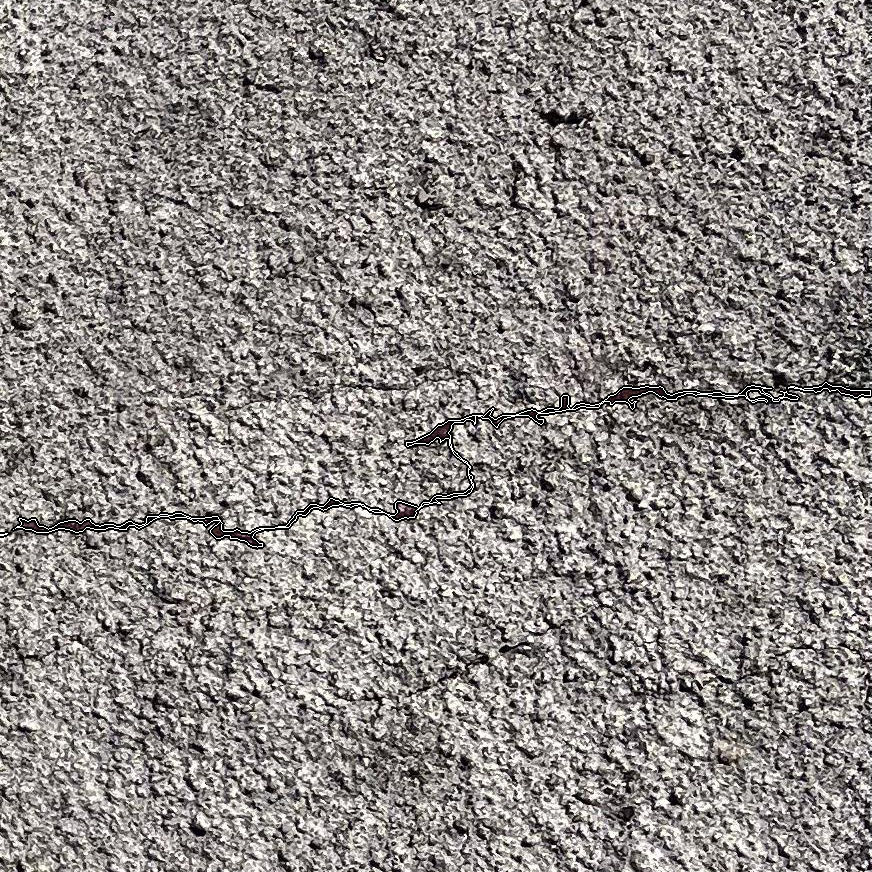}} &
                \vcenteritem{\includegraphics[width=0.14\textwidth]{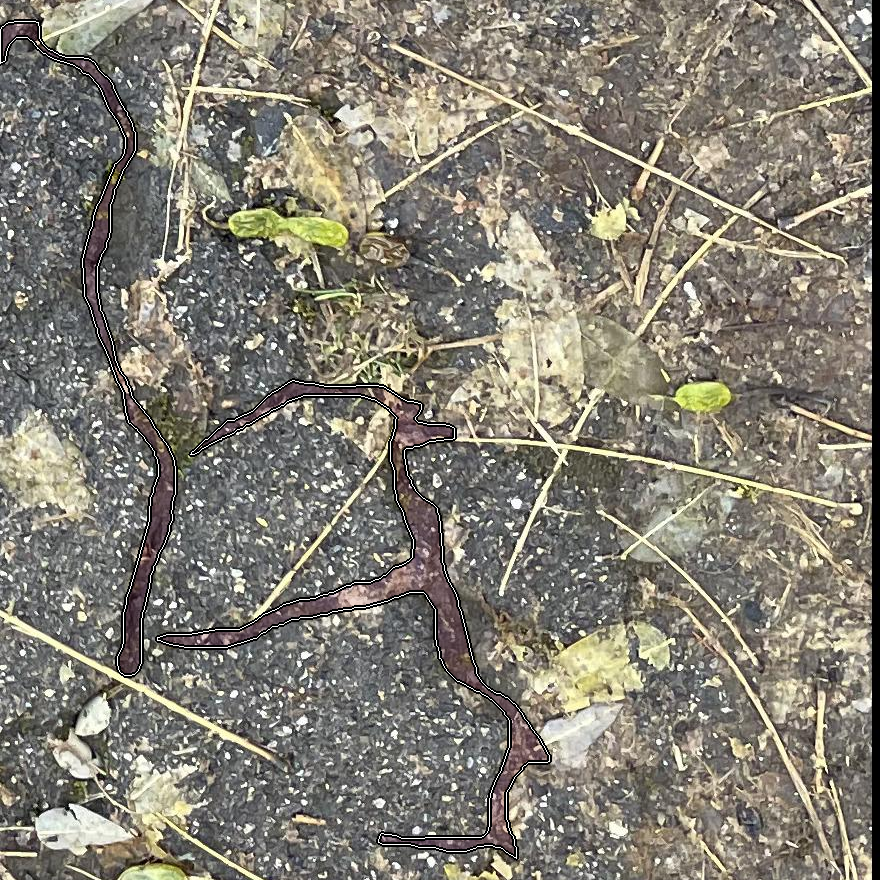}} &
                \vcenteritem{\includegraphics[width=0.14\textwidth]{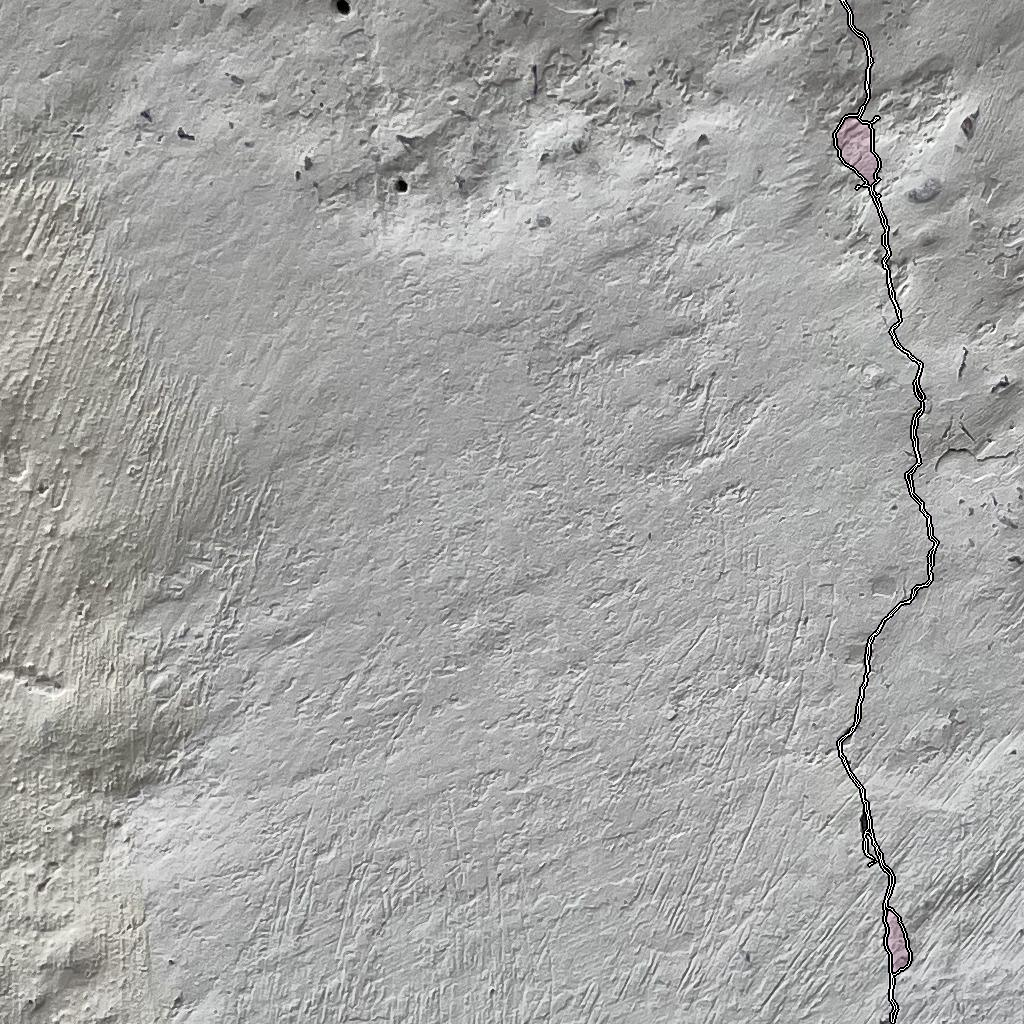}} &
                \vcenteritem{\rotatebox[origin=c]{-90}{Crack}} \\[10mm]

                \vcenteritem{\raisebox{5mm}{\rotatebox[origin=c]{90}{Net-crack}}} &
                \vcenteritem{\includegraphics[width=0.14\textwidth]{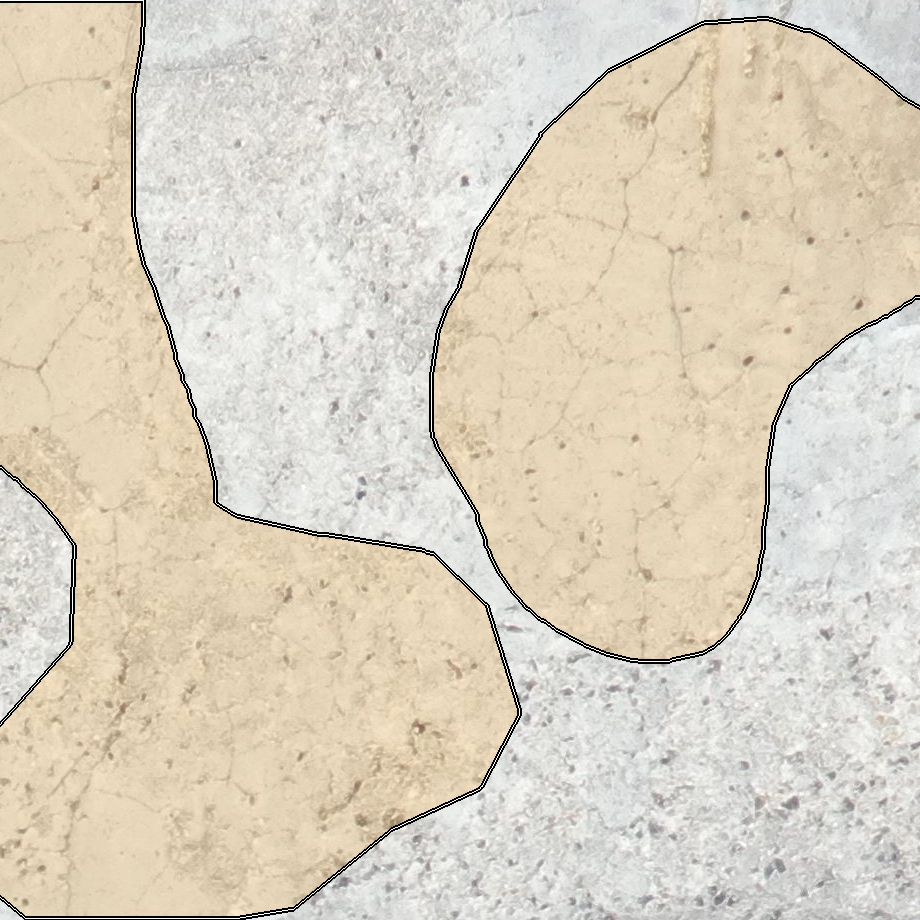}} &
                \vcenteritem{\includegraphics[width=0.14\textwidth]{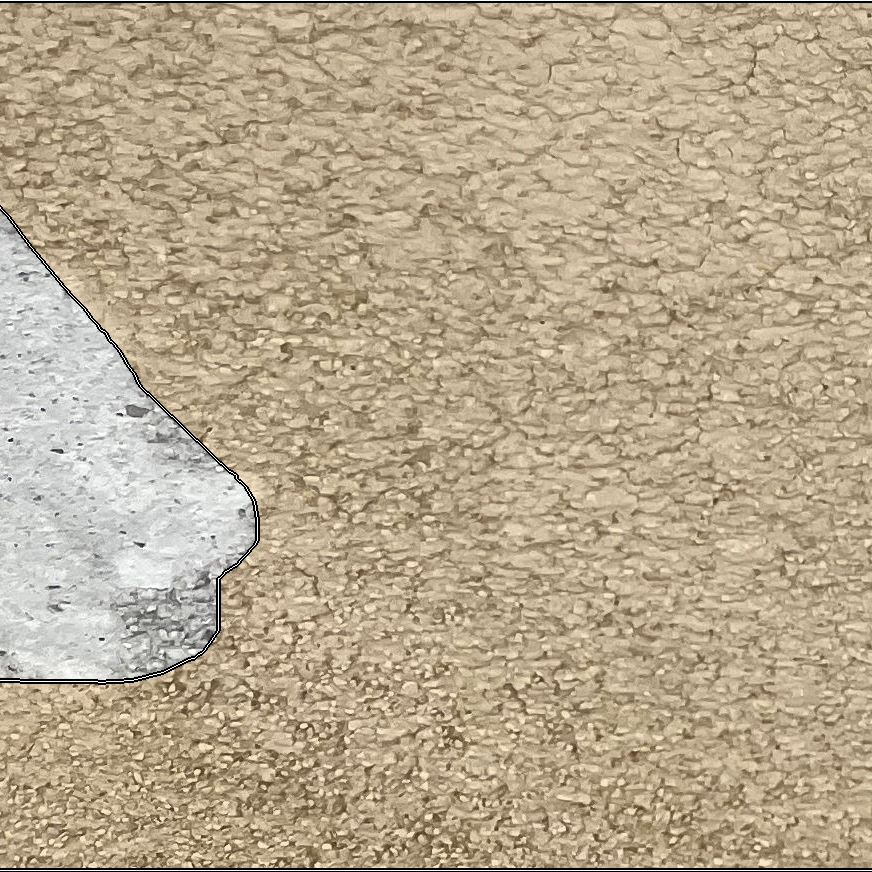}} &
                \vcenteritem{\includegraphics[width=0.14\textwidth]{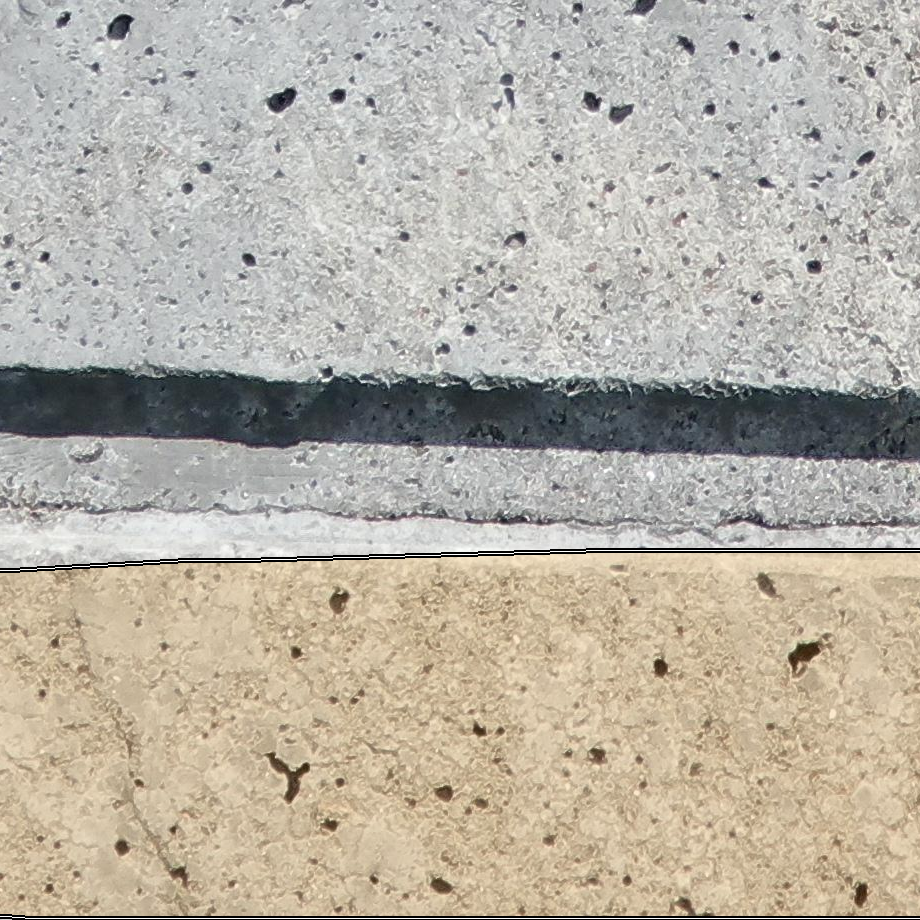}} &
                \vcenteritem{\includegraphics[width=0.14\textwidth]{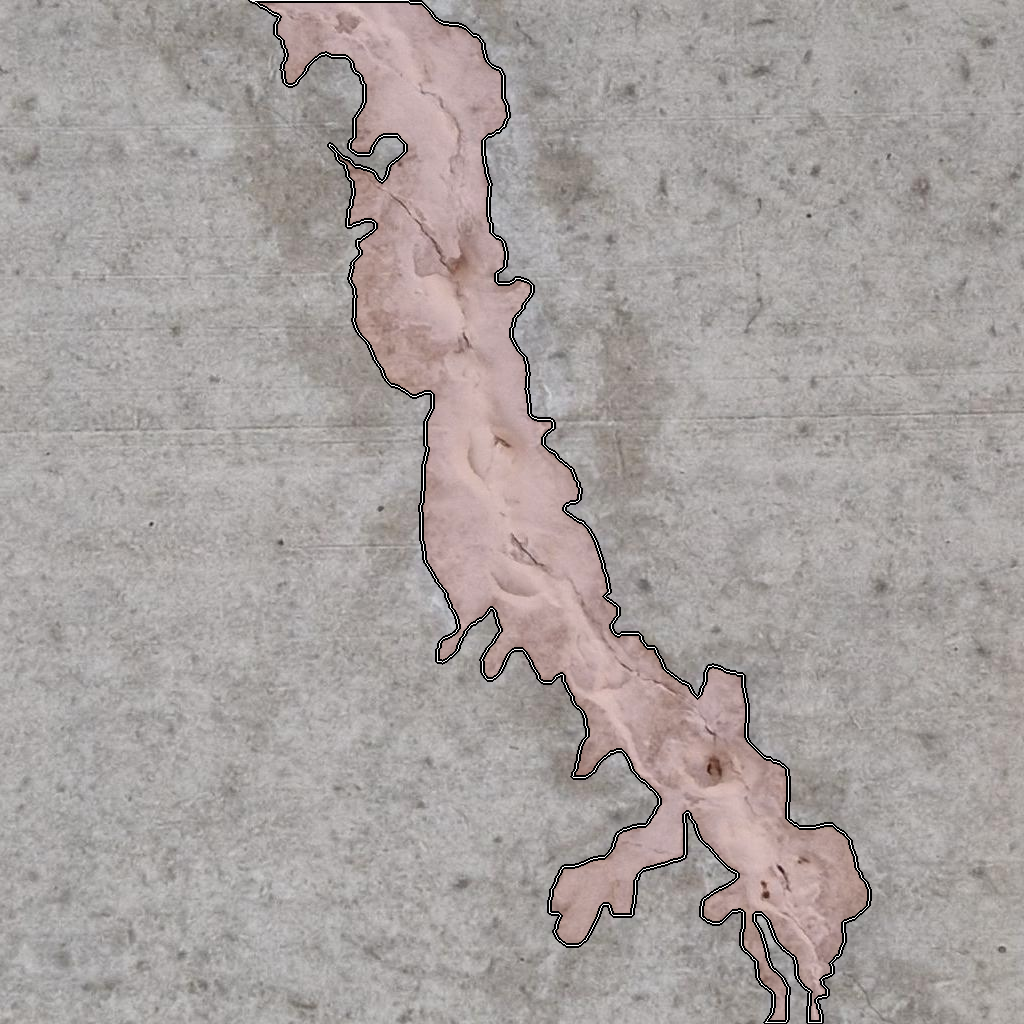}} &
                \vcenteritem{\includegraphics[width=0.14\textwidth]{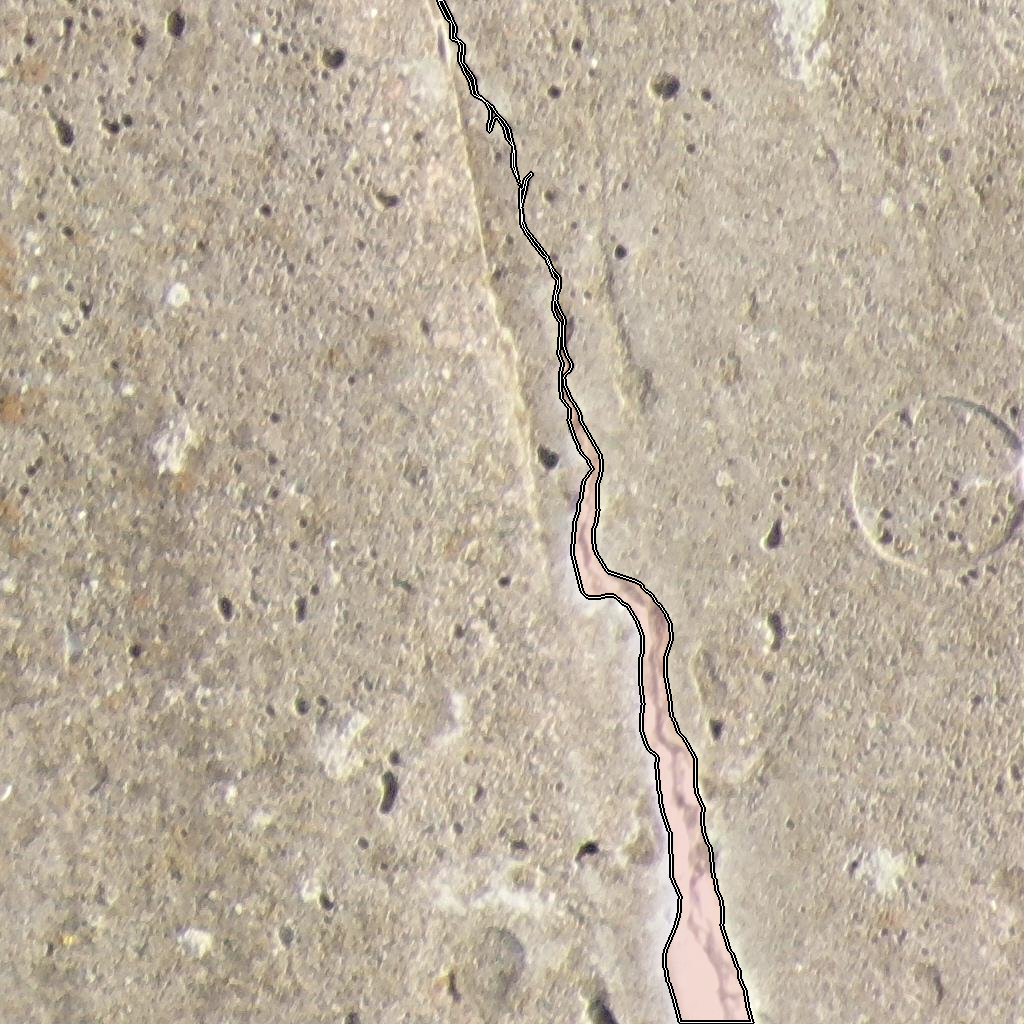}} &
                \vcenteritem{\includegraphics[width=0.14\textwidth]{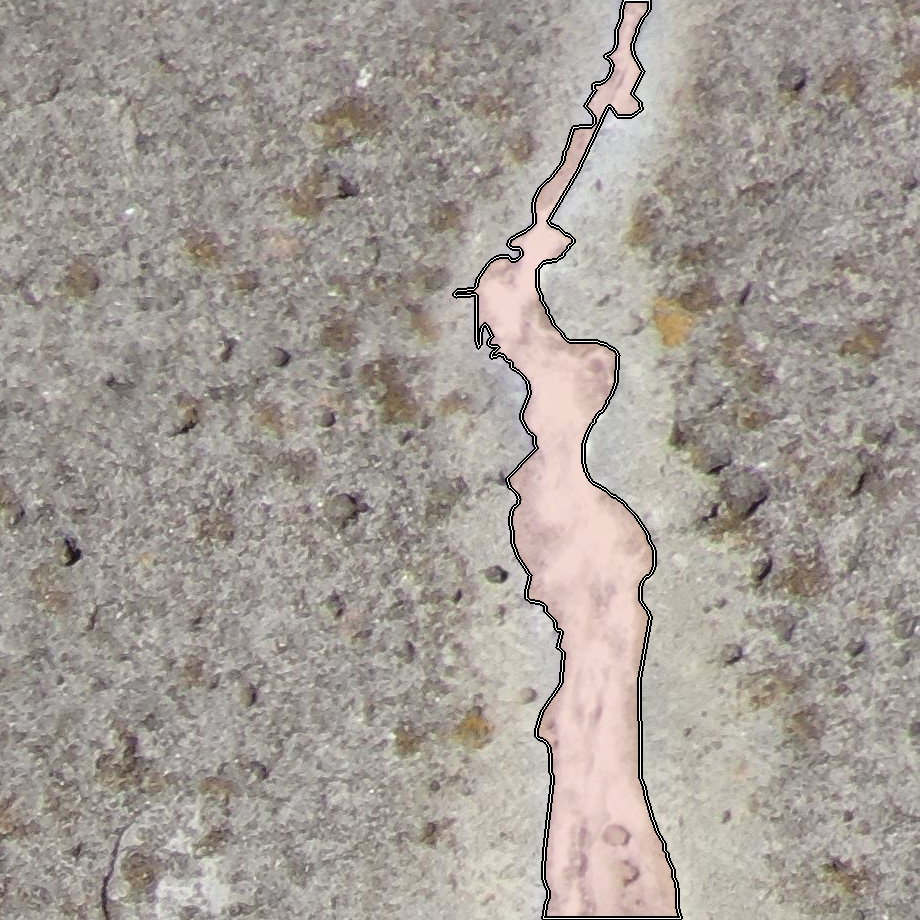}} &
                \vcenteritem{\raisebox{2mm}{\rotatebox[origin=c]{-90}{Precip}}} \\[10mm]

                \vcenteritem{\rotatebox[origin=c]{90}{Rust}} &
                \vcenteritem{\includegraphics[width=0.14\textwidth]{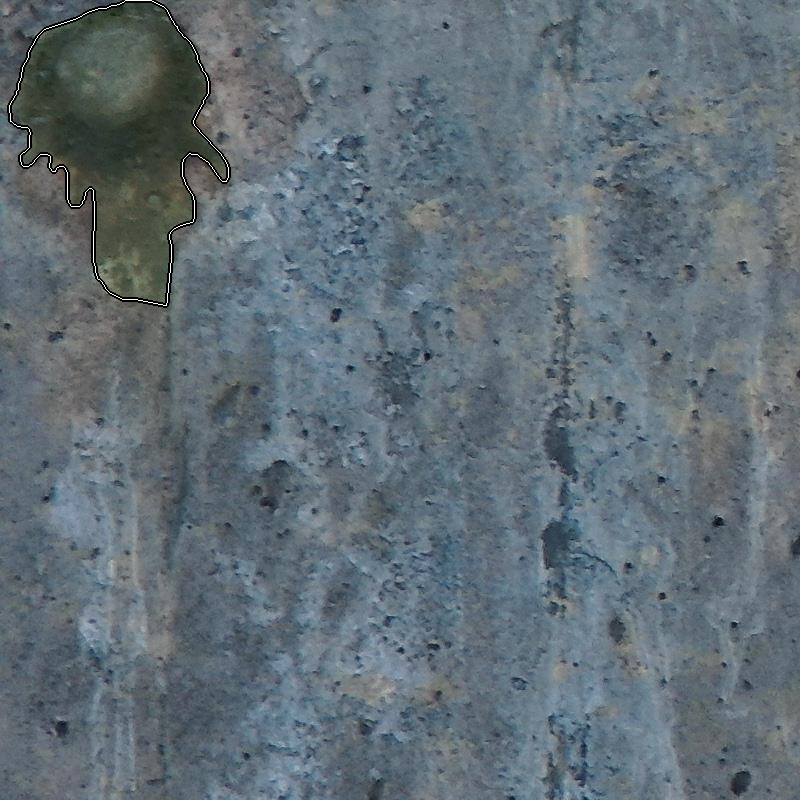}} &
                \vcenteritem{\includegraphics[width=0.14\textwidth]{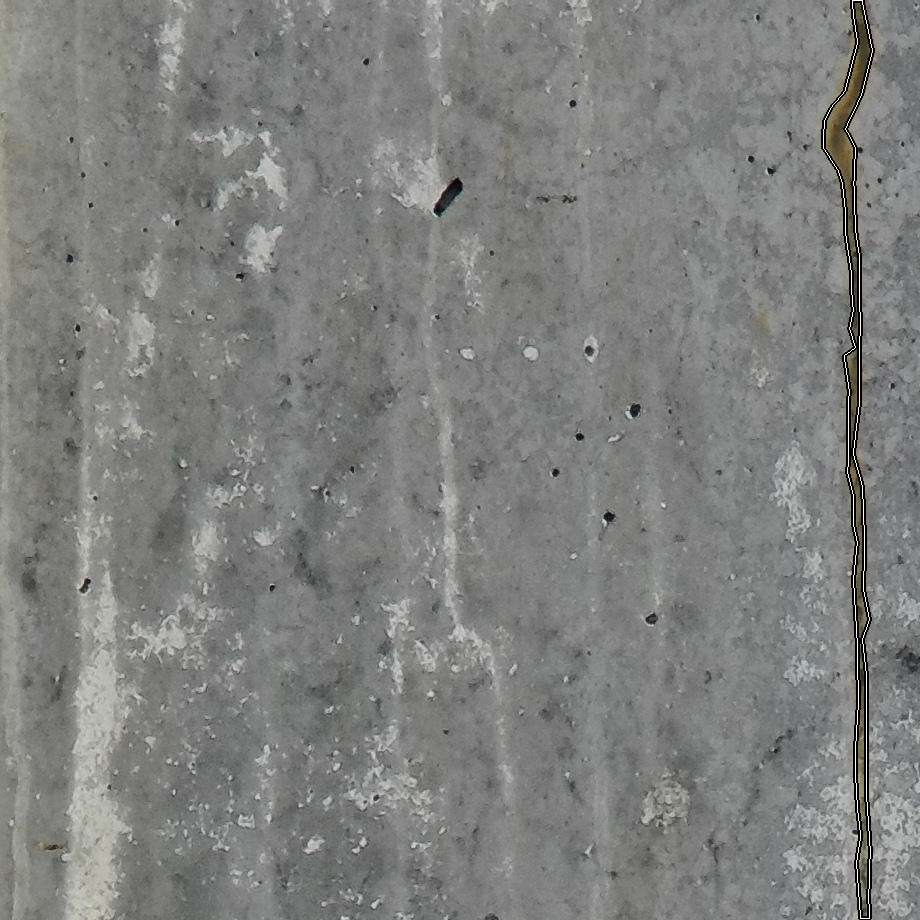}} &
                \vcenteritem{\includegraphics[width=0.14\textwidth]{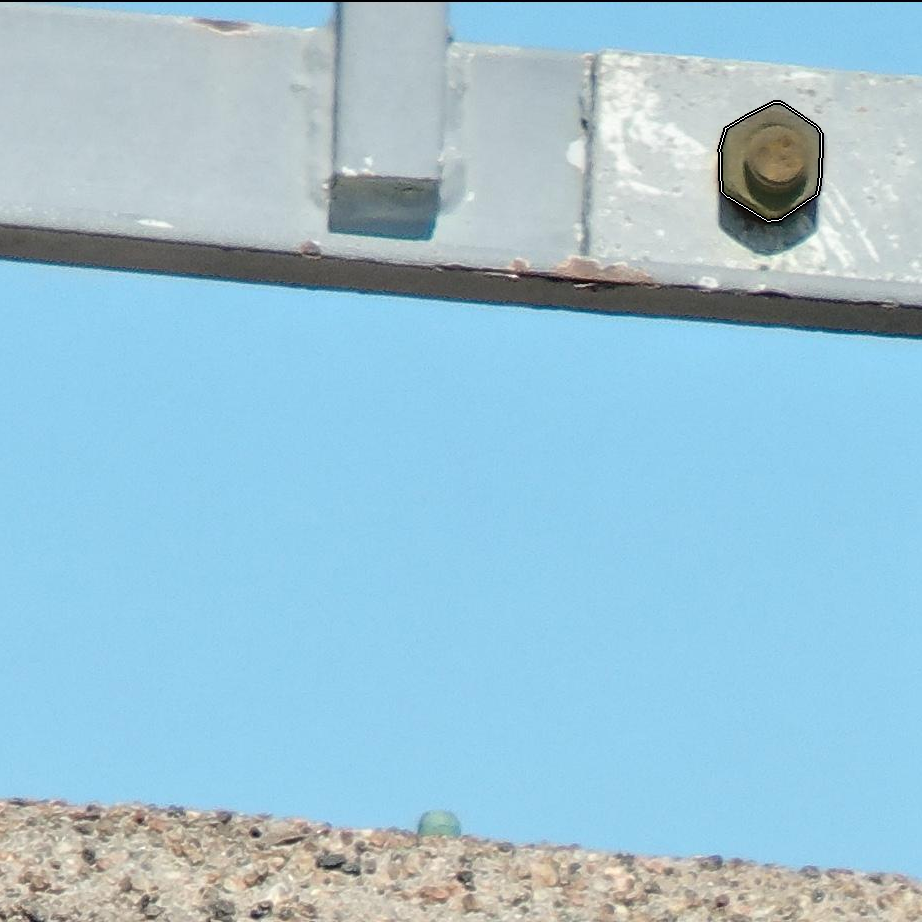}} &
                \vcenteritem{\includegraphics[width=0.14\textwidth]{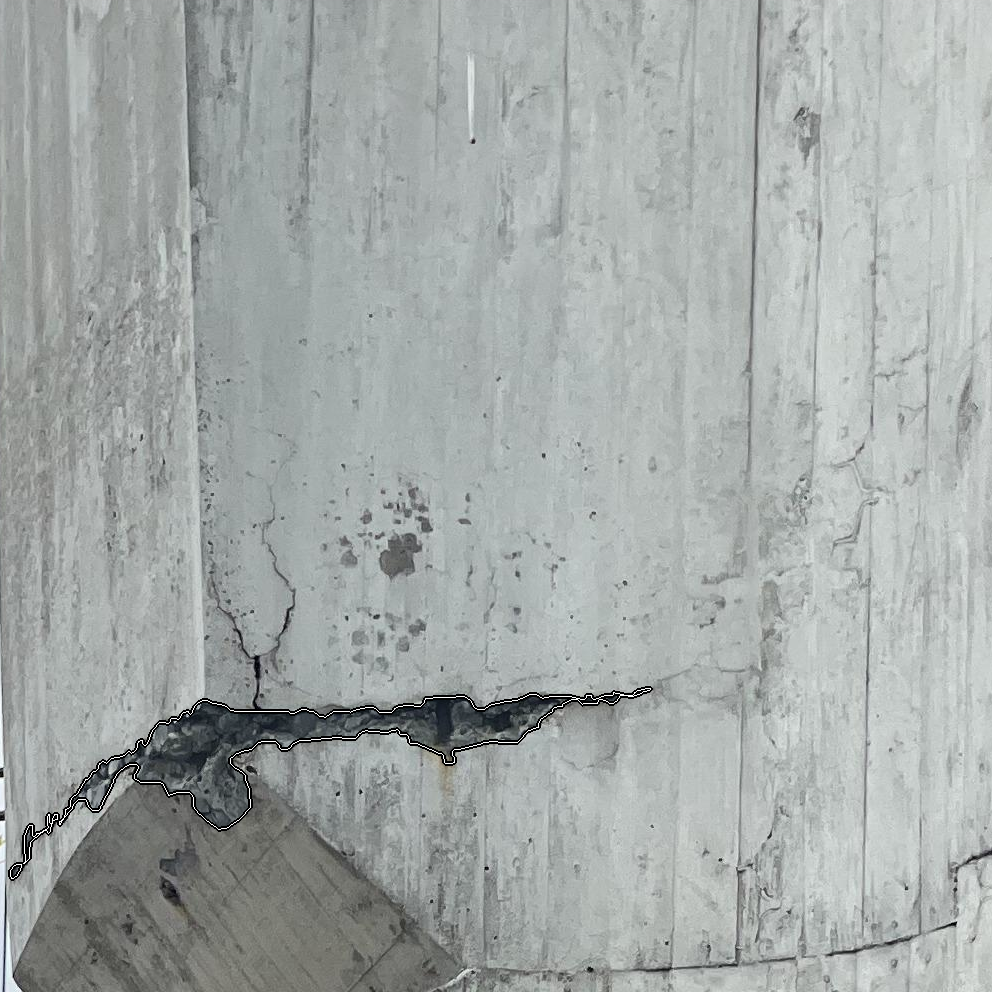}} &
                \vcenteritem{\includegraphics[width=0.14\textwidth]{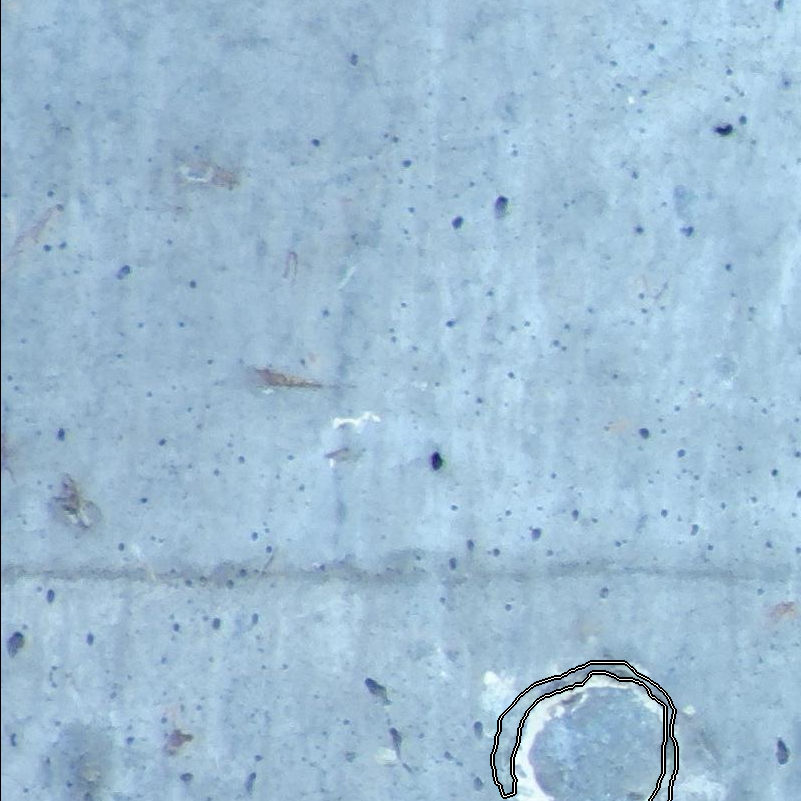}} &
                \vcenteritem{\includegraphics[width=0.14\textwidth]{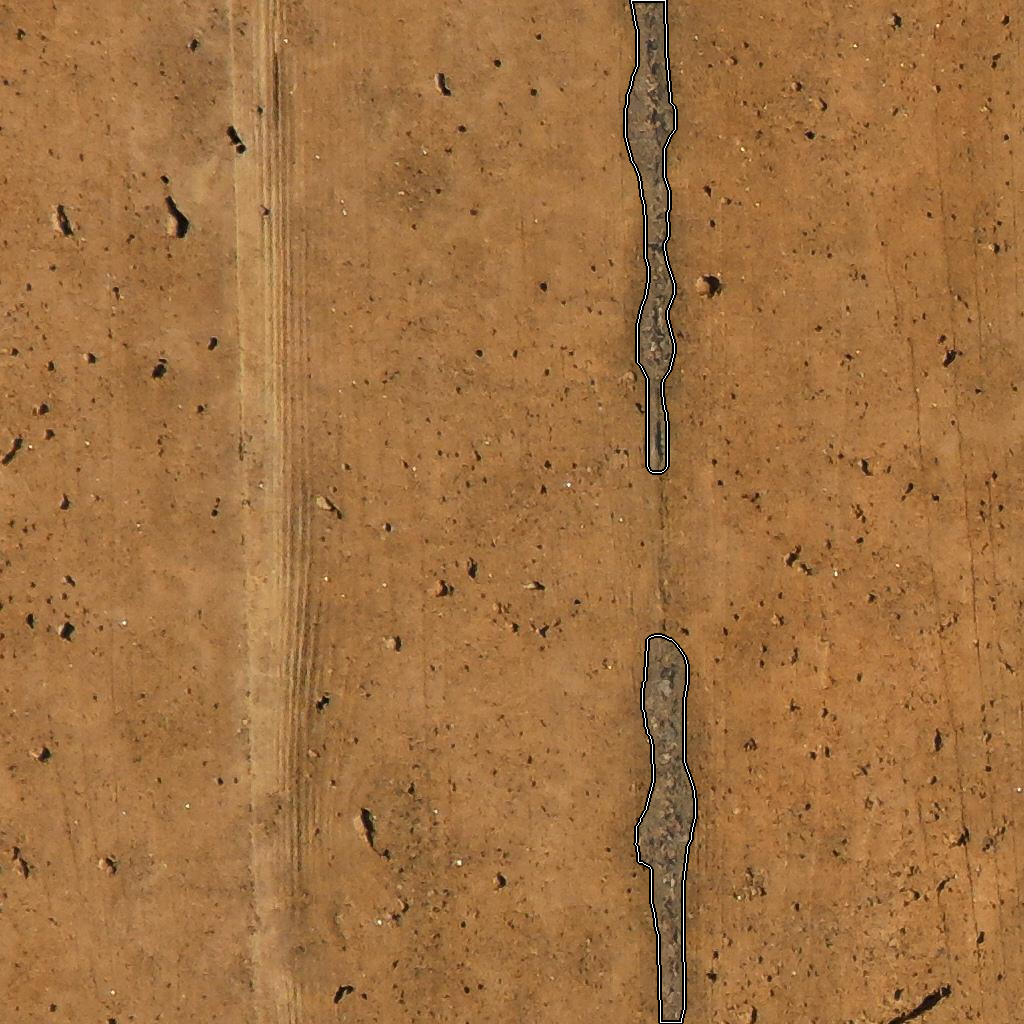}} &
                \vcenteritem{\raisebox{3mm}{\rotatebox[origin=c]{-90}{Spalling}}} \\
                
            \end{tabular}%
            }
        }; 

        \spy [halo spy=cracks_magenta, magnification=2, size=1.0cm] on (6.25, 2.5) in node at (5.2, 2);
        \spy [halo spy=cracks_magenta, magnification=2, size=0.7cm] on (2.5, 3) in node at (3.75, 2.5);
        \spy [halo spy=cracks_magenta, magnification=4, size=0.7cm] on (0.5, 2) in node at (1.5, 2.75);

    \end{tikzpicture}
    
    \caption{Mosaic of the six defect types in tiled images.}
    \label{fig:defects_mosaic_fullwidth}
\end{figure}

\paragraph{Classes.}
The six categories covering the dominant visual defect families on aged concrete and steel surfaces are illustrated in Figure~\ref{fig:defects_mosaic_fullwidth}.
\emph{Algae}: biological surface growth indicating sustained moisture.
\emph{Crack}: linear discontinuities in the concrete or coating.
\emph{Net-Crack}: interconnected polygonal cracking patterns (alligator or map cracking) associated with fatigue or shrinkage.
\emph{Crack with Precipitation}: crack with mineral deposits left by water seepage (efflorescence).
\emph{Rust}: oxidation of exposed or embedded steel reinforcement.
\emph{Spalling}: detachment or fragmentation of surface material.

\paragraph{Splits.}
Images are partitioned into train (70\%), validation (15\%), and test (15\%) at the annotation batch level. Each batch corresponds to an inspection or a defined subset of one, and tiles inherit their parent's split, so no data leaks between splits across resolutions. To prevent spatial leakage, data is organized by owner, asset, and inspection batch; while the same bridge may appear across splits, exact physical locations never overlap because separate inspections target distinct parts of the structure. This methodology yields 12{,}896 Full images (9{,}096 train, 1{,}942 val, 1{,}858 test) and 148{,}642 Tiled patches (105{,}139 train, 22{,}368 val, 21{,}135 test), all of which are fully open-sourced alongside the official evaluation splits.

\subsection{Annotation Procedure}
\label{sec:annotation_procedure}
All defects are annotated as instance-level polygon masks at native image resolution, capturing both hairline cracks a few pixels wide and large-extent defects spanning a substantial fraction of multi-megapixel frames. The class taxonomy of \S\ref{sec:dataset_details} was finalised in close collaboration with civil engineers at Sund \& B\ae{}lt, restricted to six categories that are simultaneously severe enough to warrant pixel-level localisation and operationally actionable for structural inspection and maintenance. Polygon mask annotation was carried out by an external annotation vendor under written guidelines authored jointly with the domain experts, with quality assurance performed both internally by the vendor and by our in-house reviewers.

\paragraph{Annotation guidelines.}
A written guideline document was produced for the annotation team, defining each defect type by a textual description, positive examples, and characteristic false positives. Because defect appearance, scale, and confusable background structures vary markedly across acquisition modalities and asset classes, separate guideline variants were issued per image source (drone, phone camera, DSLR) and per structure type (concrete bridge, city infrastructure, etc.). The guidelines were maintained as a living document throughout the five-year campaign: ambiguous samples surfaced by the annotation team were triaged with the domain experts and, once resolved, folded back as new canonical examples. This loop is the primary mechanism by which we controlled the dominant source of label noise in this domain---defect ambiguity at the boundary between visually similar categories. Our segmentation masks strictly adhere to standard inspection practices~\cite{astm_d6433,bast}. Net-Crack and Spalling are area-based defects, so we annotate them as filled blobs. Individual Cracks are linear, so we annotate them as thin polygons and derive their width post-hoc.

\paragraph{Pre-annotation triage.}
Two classes of images were excluded prior to annotation. Frames judged to be of insufficient image quality (motion blur, severe defocus, or extreme exposure) were discarded. Frames that contained no defect of interest were treated as background and excluded from the annotated set; they do not contribute to the released splits.

\paragraph{Batches and pilot phase.}
Annotation proceeded in batches of approximately 500--1{,}000 images, grouped by image source and structure type, which we refer to as a \emph{batch type}. For every new batch type, the annotation team first delivered a pilot of 10--50 images, used both to calibrate annotators to the specific imaging conditions and structure and to surface guideline gaps before they propagated through a full batch. Pilots were reviewed end-to-end and iterated upon with the team until quality was acceptable; a full batch was released only after pilot sign-off.

\paragraph{Full-batch annotation and QA.}
Full annotation batches were produced by an external vendor, who conducted an initial internal Quality Assurance (QA) pass prior to delivery. Upon receipt, our in-house ML research team---trained by domain experts, primarily at Sund \& B\ae{}lt---executed a rigorous second-pass audit by performing an instance-level review on a representative 10\% sample of the images. If this audit revealed an instance-level false-positive or false-negative rate exceeding our strict 10\% threshold, the entire batch was rejected and returned for correction; otherwise, localized errors were rectified and the batch was approved. To establish a robust baseline, the first five batches underwent a comprehensive 100\% review. This initial bootstrapping phase served both to calibrate our internal reviewers against the annotation guidelines and to empirically validate that our 10\% sampling protocol accurately captured overall batch quality. Comprehensive details of our QA pipeline are provided in Appendix~\ref{app:qa}.
\subsection{Statistics}
\label{sec:statistics}

\begin{figure}[h]
  \centering
  \begin{subfigure}[t]{0.58\linewidth}
    \includegraphics[width=\linewidth]{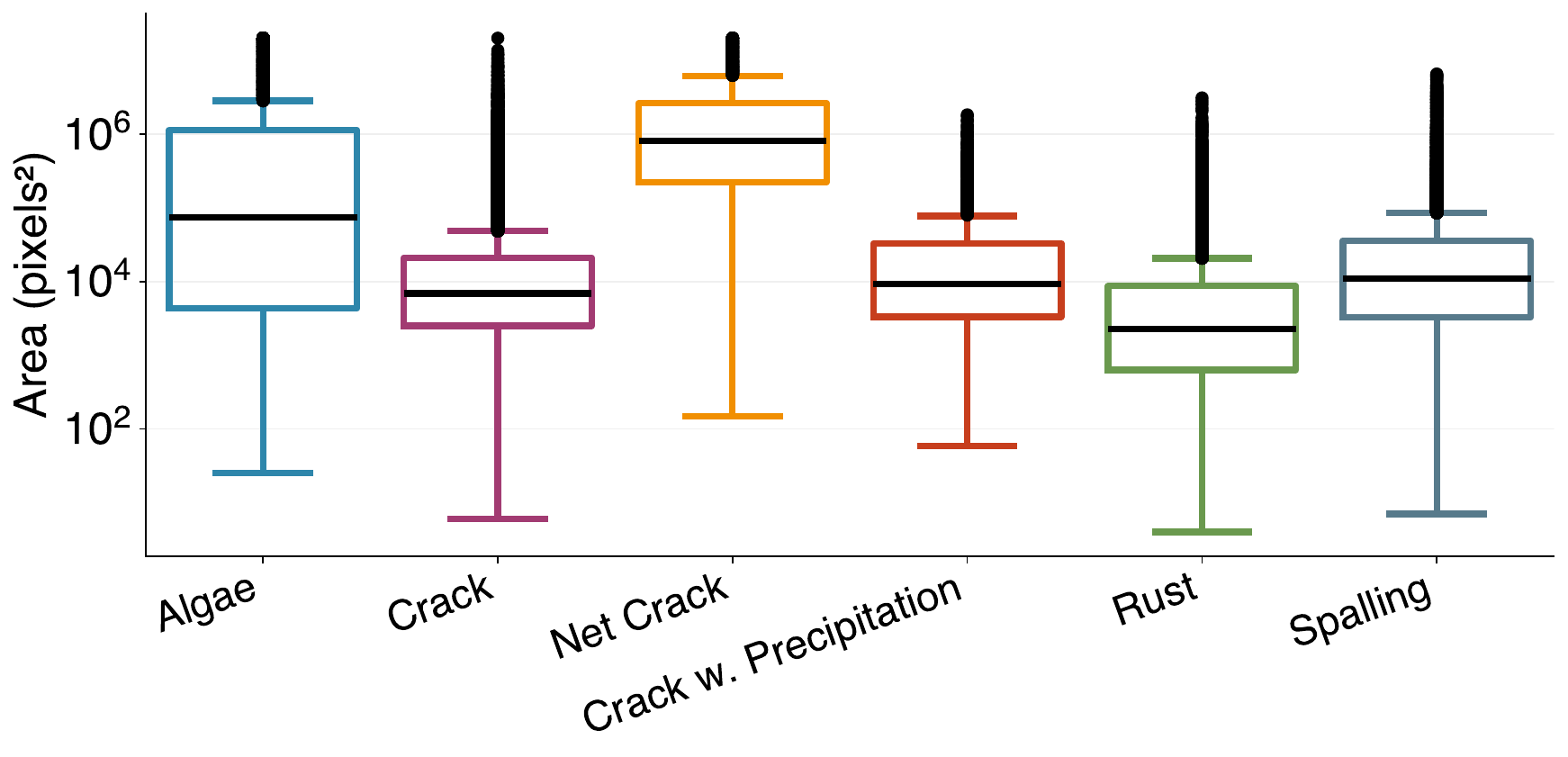}
    \caption{Defect instance area per class.}
  \end{subfigure}
  \hfill
  \begin{subfigure}[t]{0.38\linewidth}
    \includegraphics[width=\linewidth]{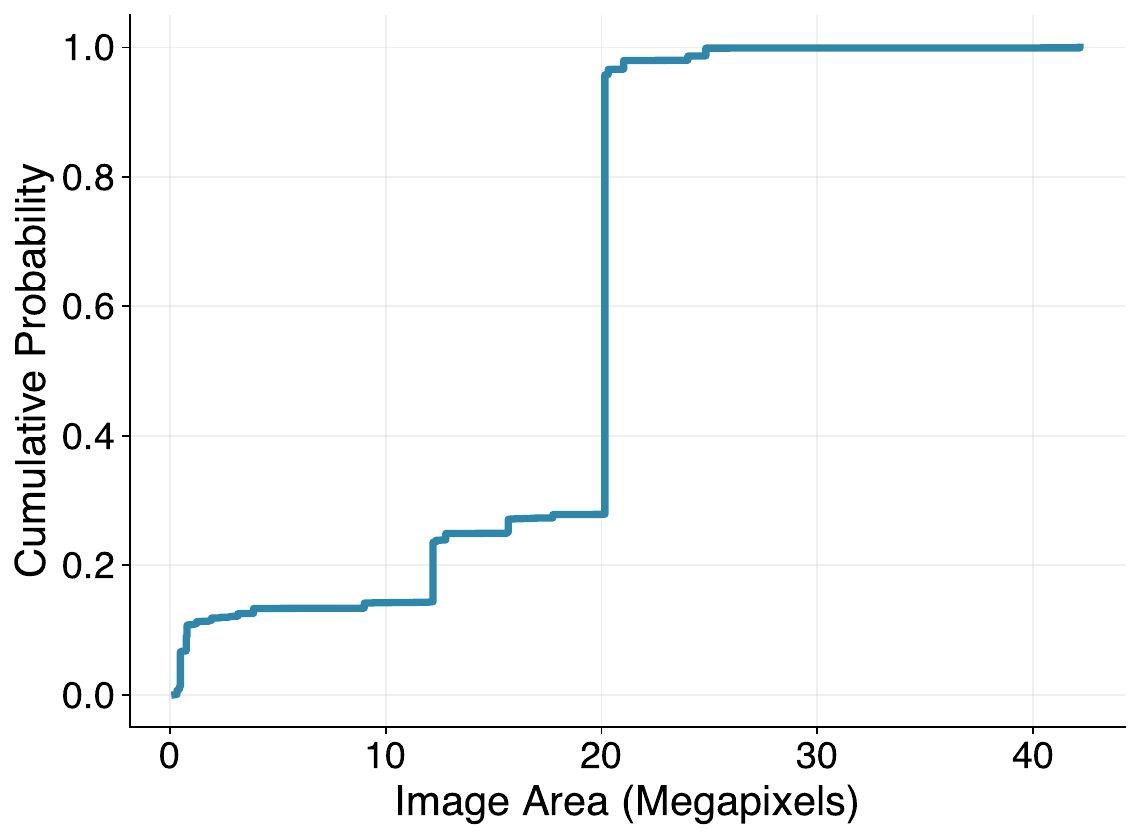}
    \caption{Cumulative image resolution distribution.}
  \end{subfigure}
  \caption{Quantitative properties of the Full variant: distribution of defect areas across classes (a) and cumulative native image resolution distribution (b).}
  \label{fig:scale_resolution}
\end{figure}

The \emph{Full} variant is characterized by its high native resolution, which significantly exceeds that of standard generic instance segmentation benchmarks. With a mean of 16.66\,MPx and a median of 20.16\,MPx, the images are approximately 50$\times$ larger than COCO ($\sim$0.3\,MPx) \cite{lin2014microsoft} and 8$\times$ larger than Cityscapes (2.1\,MPx) \cite{cordts2016cityscapes}. The cumulative distribution shown in Fig.~\ref{fig:scale_resolution}b reveals that approximately 70\% of the images have a resolution of at least 20\,MPx, reflecting the raw output of operational field hardware: smartphones, DSLRs, and gimbal-mounted survey cameras. Such high resolution is essential to capture hairline defects and high-frequency structural textures that would be lost in standard downsampled benchmarks. 
\begin{figure}[h]
  \centering
  \begin{subfigure}[t]{0.48\linewidth}
    \includegraphics[width=\linewidth]{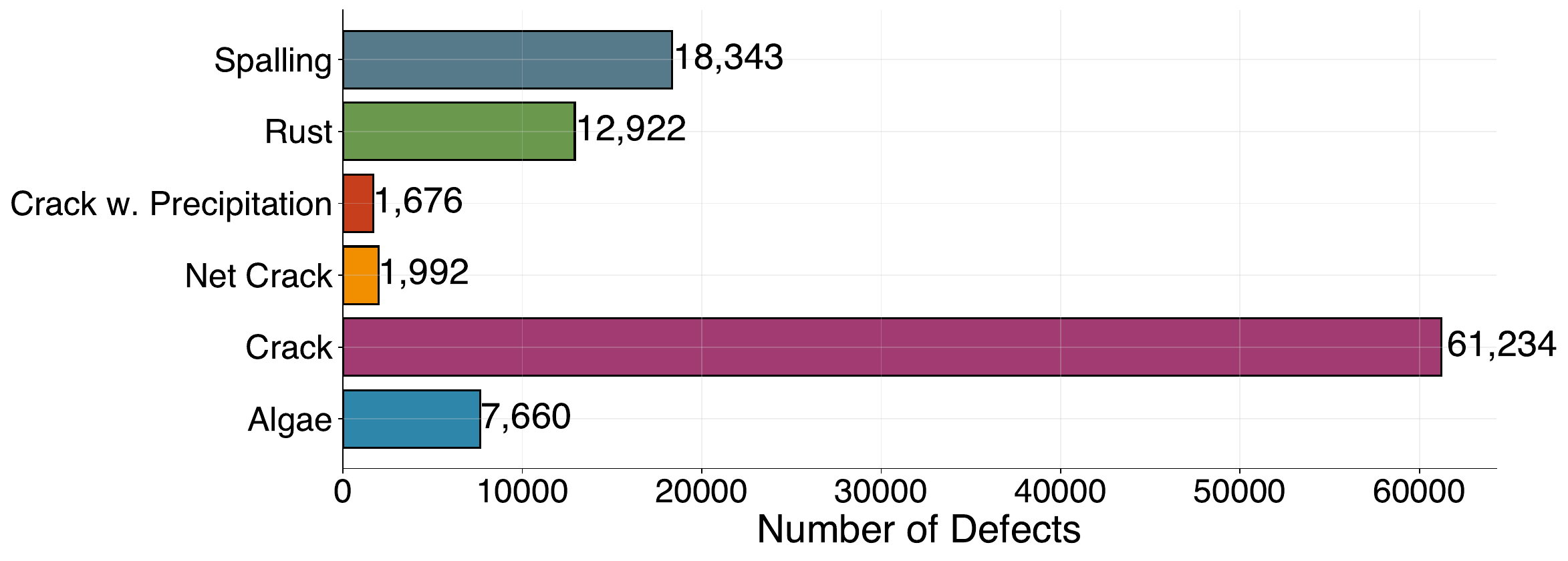}
    \caption{Full variant.}
  \end{subfigure}
  \hfill
  \begin{subfigure}[t]{0.48\linewidth}
    \includegraphics[width=\linewidth]{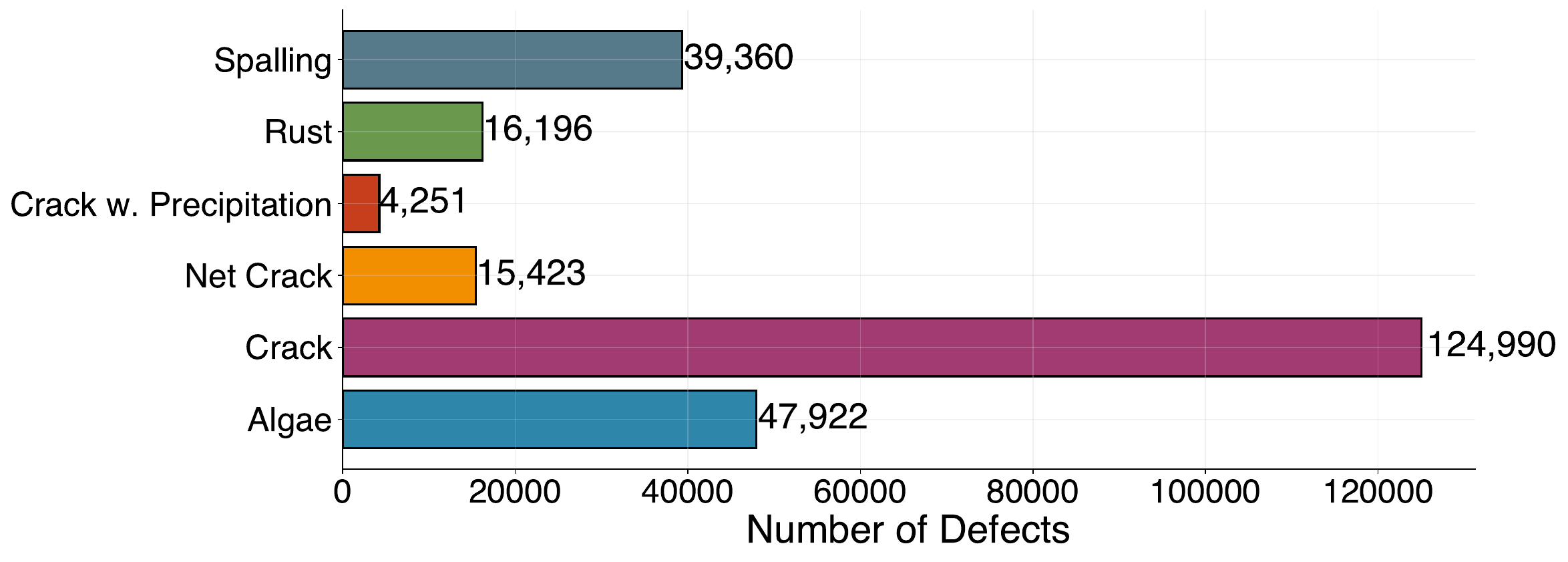}
    \caption{Tiled variant.}
  \end{subfigure}
  \caption{Per-class defect instance counts in the Full and Tiled variants.}
  \label{fig:class_dist}
\end{figure}

\paragraph{Instance Density and Class Distribution.}
The dataset contains a total of 103,827 defect instances in the Full variant. On average, each image contains 8.05 instances, a density that is higher than both generic benchmarks like COCO \cite{lin2014microsoft} ($\sim$7.3 instances/image) and specialized infrastructure benchmarks like CODEBRIM \cite{CODEBRIM} ($\sim$3.37 instances/image). 
The class distribution is strongly long-tailed (Fig.~\ref{fig:class_dist}). In the Full variant, Crack accounts for 59.0\% of all instances, while Spalling and Rust represent 17.7\% and 12.4\% respectively. Rarer categories such as Net-Crack (1.9\%) and Crack with Precipitation (1.6\%) still provide nearly two thousand instances each (Net-Crack: 1{,}992, Precipitation: 1{,}676 in Full). The Tiled variant exhibits a different distribution due to the partitioning of spatially extensive defects: Algae rises to 19.3\% of instances, while Crack and Spalling account for 50.4\% and 15.9\% respectively. We preserve the natural frequency of defect classes without artificially resampling or balancing the data. This ensures the dataset's statistical prior accurately reflects real-world conditions, where cracks naturally dominate as the primary physical precursor to secondary deterioration such as rust and spalling~\cite{structural_concrete_review}.

\paragraph{Defect Scale and Geometric Diversity.}
Beyond image scale, the difficulty profile is dominated by the wide span of defect instance areas across and within classes. Median instance areas span over two orders of magnitude across categories: from $\sim$2.3\,K\,px$^{2}$ for Rust to $\sim$807\,K\,px$^{2}$ for Net-Crack. Intra-class spread is exceptionally large, with Crack instances ranging over six orders of magnitude in area (Fig.~\ref{fig:scale_resolution}a). While the Full variant preserves these extremes, the Tiled variant filters out small defects (minimum area $\sim$0.5\,K\,px$^2$) and partitions large defects (e.g., Algae and Net-Crack) into many tiles. Furthermore, the defects exhibit high geometric diversity. While Cracks are characteristically elongated (mean aspect ratio 4.4), Algae and Net-Cracks often cover irregular, spatially extensive areas with higher mean compactness (0.67 and 0.85, respectively, in Full). This structural variety prevents models from relying on simple shape priors and requires robust spatial reasoning across multiple scales.

Baselines in this paper operate on the Tiled variant; additional experiments on the Full variant are reported in Appendix~\ref{sec:full_res}; the Full variant is released to support evaluation of resolution-aware and tile-aggregation methods. Class co-occurrence and per-class shape compactness are reported in Appendices~\ref{sec:appendix_cooccurrence}, and~\ref{sec:appendix_shape}.

\section{Evaluation}
\label{sec:experiments}

\begin{table*}
\caption{Object detection (bb) and instance segmentation (seg) results for supervised baselines on the CiF dataset. Models were trained on the training set, and the best-performing checkpoint on the validation set was used for final evaluation on the test set.}
\label{tab:combined-results}
\centering
\setlength{\tabcolsep}{1.2mm}
\resizebox{\linewidth}{!}{
\begin{tabular}{ll c c @{\hskip 5mm} c c c c c c c c c c c c}
\toprule
\multirow{3}{*}{Method} & \multirow{3}{*}{Encoder} & \multicolumn{2}{c}{\multirow{2}{*}{$mAP_{50:95}$}} & \multicolumn{12}{c}{$AP_{50:95}$} \\
\cmidrule(lr){5-16}
& & \multicolumn{2}{c}{} & \multicolumn{2}{c}{Algae} & \multicolumn{2}{c}{Crack} & \multicolumn{2}{c}{Net-Crack} & \multicolumn{2}{c}{Precipitation} & \multicolumn{2}{c}{Rust} & \multicolumn{2}{c}{Spalling} \\
\cmidrule(lr){3-4} \cmidrule(lr){5-6} \cmidrule(lr){7-8} \cmidrule(lr){9-10} \cmidrule(lr){11-12} \cmidrule(lr){13-14} \cmidrule(lr){15-16}
& & bb & seg & bb & seg & bb & seg & bb & seg & bb & seg & bb & seg & bb & seg \\
\midrule
Mask-DINO~\cite{li2023maskdino} & Resnet-50 & 29.0 & 20.4 & 51.2 & 46.5  & 31.3 & 3.0 & 27.2 & 22.8 & 14.1 & 4.3 & 26.9 & 26.7 & 23.4 & 19.0 \\
\midrule
\multirow{5}{*}{YOLOv11~\cite{khanam2024yolov11}} 
 & nano   & 31.0 & 19.8 & 52.6 & 43.6 & 31.0 & 6.5 & 29.0 & 21.0 & 17.1 & 4.9 & 30.0 & 25.1 & 26.1 & 17.7 \\
 & small  & 32.3 & 20.7 & 53.0 & 43.9 & 32.9 & 7.5 & 29.2 & 21.7 & 18.3 & 5.9 & 32.8 & 26.7 & 27.4 & 18.7 \\
 & medium & 33.2 & 21.5 & 53.4 & 44.3 & 34.6 & 8.2 & 30.2 & 23.2 & 18.8 & 6.3 & 34.0 & 27.4 & 28.4 & 19.7 \\
 & large  & 33.9 & 21.9 & 53.8 & 44.7 & 35.6 & 8.5 & 30.8 & 23.5 & 19.9 & 6.7 & 34.7 & 27.8 & 29.2 & 20.3 \\
 & xlarge & 33.9 & 21.9 & 53.6 & 44.3 & 35.6 & 8.6 & 30.7 & 23.4 & 19.7 & 7.2 & 35.0 & 28.2 & 28.7 & 19.9 \\
\midrule
\multirow{5}{*}{YOLOv26~\cite{sapkota2025yolo26}} 
 & nano   & 30.2 & 19.5 & 52.0 & 43.5 & 30.3 & 6.8 & 27.8 & 20.7 & 17.4 & 4.9 & 28.2 & 23.6 & 25.2 & 17.4 \\
 & small  & 32.2 & 20.9 & 52.7 & 43.9 & 32.5 & 7.6 & 29.1 & 21.8 & 19.6 & 7.0 & 31.7 & 25.9 & 27.6 & 19.0 \\
 & medium & 32.8 & 21.4 & 53.0 & 44.3 & 33.2 & 8.1 & 29.8 & 22.9 & 20.2 & 6.9 & 33.3 & 27.3 & 27.3 & 19.1 \\
 & large  & 32.0 & 20.2 & 51.7 & 42.8 & 33.1 & 7.9 & 27.5 & 20.4 & 20.1 & 6.3 & 32.4 & 25.6 & 26.8 & 18.5 \\
 & xlarge & 32.7 & 21.0 & 52.2 & 43.4 & 34.1 & \textbf{8.9} & 27.2 & 20.6 & 20.1 & 6.8 & 33.7 & 26.2 & 28.9 & 20.3 \\
\midrule
\multirow{4}{*}{RF-DETR~\cite{rf-detr}} 
 & nano   & 35.3 & 18.3 & \textbf{59.3} & 46.1 & 37.5 & 1.2 & 30.6 & 21.5 & 18.6 & 7.3 & 36.4 & 16.7 & 29.7 & 17.1 \\
 & small  & \textbf{36.2} & 18.2 & 59.2 & 46.2 & \textbf{38.3} & 1.3 & \textbf{31.8} & 21.4 & \textbf{20.5} & 7.2 & \textbf{36.8} & 17.4 & \textbf{30.5} & 15.8 \\
 & medium & 35.7 & 18.3 & 59.1 & 46.8 & 38.1 & 1.2 & 31.5 & 21.5 & 19.6 & 6.8 & 36.4 & 17.0 & 29.7 & 16.7 \\
 & large  & 35.7 & 18.3 & 58.9 & 46.5 & 38.2 & 1.3 & 31.5 & 21.0 & 19.5 & 7.6 & 36.5 & 16.7 & 29.8 & 16.6 \\
\midrule
\multirow{2}{*}{EoMT~\cite{kerssies2025eomt}} 
 & DinoV3 & 32.9 & \textbf{24.2} & 57.0 & \textbf{49.6} & 28.4 & 4.7 & 30.4 & \textbf{24.1} & 17.9 & \textbf{10.3} & 34.8 & 31.6 & 29.2 & 25.1 \\
 & DinoV2 & 32.7 & 24.1 & 56.4 & 48.8 & 29.0 & 5.0 & 29.3 & 23.5 & 16.8 & 9.5 & 35.2 & \textbf{32.6} & 29.6 & \textbf{25.2} \\

\bottomrule
\end{tabular}
}

\end{table*}

\paragraph{Supervised baselines.} To establish rigorous baselines on the CiF dataset, we evaluate a diverse suite of state-of-the-art architectures renowned for top-tier performance on standard instance segmentation benchmarks like COCO. Our selection strategically covers three distinct paradigms: (i) Industry Standards: We include the YOLO family (specifically YOLOv11 \cite{khanam2024yolov11} and YOLOv26 \cite{sapkota2025yolo26}) as the de facto standards for real-world, high-performance segmentation; (ii) DETR-Based Architectures: We evaluate Mask DINO~\cite{li2023maskdino} as a strong DETR-based~\cite{carion2020end} segmentation and detection baseline, alongside RF-DETR \cite{rf-detr}, a state-of-the-art compact and computationally efficient transformer model; and (iii) Pretrained Foundations: We include EoMT \cite{kerssies2025eomt} to assess knowledge transfer of strong, pretrained foundational encoders to the civil infrastructure domain. Specifically, we aim to probe the out-of-domain finetuning of foundation backbones like DINOv2/v3 \citep{oquab2024dinov2learningrobustvisual, simeoni2025dinov3} using a state-of-the-art algorithm like EoMT. All baselines are fully fine-tuned on the CiF training split, starting from standard pretrained weights if available with optimal checkpoints selected via validation performance before final evaluation on the held-out test set. The results are detailed in Table~\ref{tab:combined-results}, reporting the $\text{mAP}_{50:95}$ performance for both the instance segmentation masks and the bounding box predictions. More details about the training setups can be found in Appendix~\ref{sec:appendix_training_details}.

\paragraph{Foundation Model Evaluation.} Given the recent success of Foundation Models (FMs) and Vision-Language Models (VLMs) claiming universal, zero-shot detection and segmentation capabilities~\cite{kirillov2023segment, lai2024lisareasoningsegmentationlarge, bai2025qwen3vltechnicalreport}, we conduct a rigorous evaluation of these architectures on the CiF dataset. We specifically investigate two dominant paradigms: promptable segmentation and zero-shot object detection. For promptable segmentation, we evaluate SAM3 \citep{carion2026sam3segmentconcepts}, the latest iteration of the Segment Anything family \citep{kirillov2023segment, ravi2024sam2segmentimages}. To thoroughly assess its adaptability, we evaluate SAM3 under three distinct prompting protocols: oracle bounding boxes derived from ground truth, oracle point-based prompting (randomly sampling 1 to 5 points from the ground truth mask), and text-based prompting. Because these oracle protocols rely on privileged ground-truth spatial prompts, they are designed strictly to evaluate SAM3's viability as a semi-automated annotation assistant rather than as an off-the-shelf segmenter. For zero-shot detection, we evaluate Qwen3-VL, a state-of-the-art VLM known for exceptional performance on referring expression segmentation benchmarks (e.g., refCOCO, refCOCO+ \cite{yu2016modelingcontextreferringexpressions}) and diverse vision-language tasks. Qwen3-VL is evaluated using standard off-the-shelf text prompts, instructed to output object detections in a structured JSON format (Prompt details can be found in Appendix~\ref{sec:appendix_prompt}). Moreover, we benchmark SAM3-Agent, a hybrid architecture integrating the zero-shot reasoning capabilities of Qwen3-VL with the precise dense prediction of SAM3 within an agentic framework. Quantitative results are detailed in Table~\ref{tab:qwen3-results} for instance segmentation and bounding box detection, with both evaluated using the $\text{mAP}_{50:95}$ metric.

\subsection{Discussion}
\label{sec:discussion}
\paragraph{Moderate success of supervised baselines.} Standard supervised models achieve reasonable object detection performance, yielding an overall bounding-box mAP@50:95 of approximately 35\%. The models exhibit strong localization capabilities on visually distinct, surface-level anomalies such as algae (reaching up to 59\% mAP). However, performance degrades on critical structural defects. Complex failure modes, such as \textit{net crack} or \textit{crack with precipitation}, yield localization scores of only 20–30\%, showing all the challenges of the civil infrastructure domain. When it comes to instance segmentation, performance drops drastically across all architectures. With the exception of high-contrast algae, mask mAP plunges to single digits for fine-grained structural defects like cracks. This drop in pixel-level accuracy is particularly detrimental for civil engineering applications, where the severity of a defect is strictly measured by its exact geometric footprint. Among the segmentation baselines, EoMT achieves the highest performance by leveraging an internet-scale foundational backbone. This outcome suggests that massive-scale pretraining may offer a measurable advantage in feature extraction, though it is difficult to cleanly disentangle the impact of the foundational data from the model's inherent architecture. What remains definitively clear, however, is that the absolute performance is insufficient for real-world application. This stark reality positions CiF as a vital asset and proving ground, challenging the field to determine whether future breakthroughs in civil infrastructure will be driven by specialized architectural advancements or by even larger, high-fidelity datasets.

\begin{table*}
\caption{Performance comparison of SAM3 and Qwen3-VL on the CiF dataset. The $AP_{50:95}$ metric denotes segmentation accuracy for SAM3 and object detection accuracy for Qwen3-VL. Note that SAM3-Oracle variants utilize privileged ground-truth spatial prompts strictly to assess semi-automated labeling viability and are not directly comparable to unprivileged zero-shot baselines (SAM3-Text, SAM3-Agent, and Qwen3-VL).}
\label{tab:qwen3-results}
\centering
\setlength{\tabcolsep}{1.2mm}
\resizebox{\linewidth}{!}{
\begin{tabular}{l c @{\hskip 5mm} c c c c c c}
\toprule
\multirow{2}{*}{Method} & \multirow{2}{*}{$mAP_{50:95}$} & \multicolumn{6}{c}{$AP_{50:95}$} \\
\cmidrule(lr){3-8}
& & Algae & Crack & Net-Crack & Precipitation & Rust & Spalling \\
\midrule
SAM3-Agent             & 1.7  & 1.0  & 0.7 & 0.0  & 1.0  & 7.0  & 0.7  \\
SAM3-Text              & 2.0  & 1.0  & 0.7 & 0.0  & 1.6  & 8.8  & 0.0  \\
SAM3-Oracle \#1 point  & 7.8  & 10.4 & 0.4 & 6.7  & 3.0  & 18.1 & 8.2  \\
SAM3-Oracle \#3 points & 16.1 & 30.5 & 0.6 & 20.9 & 4.6  & 25.5 & 14.7 \\
SAM3-Oracle \#5 points & 18.6 & 33.9 & 0.8 & 25.0 & 5.8  & 28.2 & 17.9 \\
SAM3-Oracle Bounding Box & 34.2 & 56.0 & 1.9 & 54.1 & 10.4 & 46.0 & 37.1 \\
\midrule
Qwen3-VL-4B   & 1.5 & 4.7 & 2.4 & 0.1 & 0.1 & 1.4 & 0.5 \\
Qwen3-VL-8B   & 1.9 & 5.8 & 2.0 & 0.1 & 0.1 & 2.8 & 0.4 \\
Qwen3-VL-32B  & 2.0 & 6.7 & 2.8 & 0.4 & 0.1 & 1.3 & 0.6 \\
Qwen3-VL-235B-A22B & 2.4 & 7.3 & 3.3 & 0.7 & 0.1 & 2.3 & 0.8 \\
\bottomrule
\end{tabular}
}

\end{table*}

\paragraph{The Illusion of Zero-Shot Generalization.} Our evaluation of Foundation Models and VLMs reveals poor zero-shot performance, exposing a critical gap between benchmark success and real-world utility. For instance, despite its massive scale, the Qwen3-VL (235B-A22B) model fundamentally fails at zero-shot object detection on civil infrastructure, yielding an abysmal mean bounding box mAP of just 2.4\%. SAM3 fares no better in zero-shot dense prediction. Even when evaluated under oracle conditions (using ground-truth bounding boxes or point prompts), performance remains strikingly low except for large defects. The model struggles with the fine-grained predictions required for the ``crack'' class, confirming that generic segmentation foundations lack the specific priors necessary for defect segmentation. Consequently, these limitations undermine SAM3's viability as an automated annotation assistant, further reinforcing the extreme difficulty and expense of curating high-fidelity training data in this domain.

\paragraph{Cracks in the Foundation.} The evaluation experiments point to an overlooked, yet vital issue in current ML literature: models that achieve state-of-the-art results on curated, general-purpose datasets consistently fail to generalize to complex, real-world domains such as civil infrastructure visual inspection. The primary objective of CiF is to spotlight this deficiency and open-source a massive-scale dataset to close this gap. Given the profound societal and economic impact of automated structural health monitoring, resolving this domain shift is not just a theoretical machine learning challenge, but a critical real-world necessity.

\subsection{Limitations}
\label{sec:limitations}
While CiF represents a large-scale and relevant dataset for civil infrastructure defect segmentation, several limitations remain. First, the dataset focuses primarily on bridge structures and concrete materials in Europe, and therefore does not fully represent the diversity of global infrastructure systems, such as steel or underground structures. Second, defect annotation is inherently subjective, particularly for fine-grained or low-contrast damage patterns, which may introduce variability even among expert annotators. Third, the data acquisition process—combining in-the-wild imagery with controlled drone-based inspections—may introduce biases in viewpoint, scale, and sensor characteristics. In addition, the dataset reflects a long-tailed distribution of defect types, which may impact model performance across rare categories.
Lastly, despite its unprecedented scale, CiF remains vastly smaller than general-purpose datasets like LAION \cite{schuhmann2022laion5bopenlargescaledataset}. While our evaluation exposes severe model failures, the root cause remains an open question. We cannot prove whether these limitations are strictly architectural; it is possible that scaling the dataset by several orders of magnitude could resolve them, though such an effort is currently economically prohibitive.

\section{Conclusions}
\label{sec:conclusions}
We introduce Cracks in the Foundation (CiF), the largest high-resolution instance segmentation dataset for civil infrastructure to date. Evaluations on CiF suggest the domain remains highly challenging: state-of-the-art Foundation Models struggle with zero-shot generalization, and specialized models plateau below real-world requirements. Ultimately, CiF establishes structural health monitoring as an open, unsolved challenge. By open-sourcing this unprecedented benchmark and large-scale training dataset, we empower the community to develop robust architectures capable of genuinely safeguarding our physical world.
\newpage

\begin{ack}
We would like to sincerely thank Finn Bormlund and Svend Gjerding (Sund \& Baelt), Jens Häggström (Trafikverket), Raphael von Thiessen (Innovation-Sandbox for AI, Office for Economy, Kanton Zürich), and the Dübendorf Air Base for granting us the opportunity to collect, analyze, and disseminate the images and defect data included in this publication. We acknowledge funding from the European Union’s Horizon Europe research and innovation programme under the Marie Skłodowska-Curie actions HORIZON-MSCA-2022-DN-01 call (Grant agreement ID: 101119554) to Nicola Farronato and Rizwan Ullah Khan, and Horizon Europe grant No. 101070408 (SustainML) to Niccolo Avogaro and Mattia Rigotti.
\end{ack}

\bibliographystyle{plain}
\bibliography{references}


\newpage
\appendix

\section{Technical Appendices and Supplementary Material}
\label{sec:appendix}
\subsection{Training Details}
\label{sec:appendix_training_details}

All baselines were trained using the off-the-shelf configurations provided in their respective official implementations. The single exception was the YOLOv26-large and YOLOv26-xlarge models, where the default learning rate caused training divergence; consequently, we scaled the learning rate down by an order of magnitude. We consciously refrained from exhaustive hyperparameter optimization. While extensive tuning might yield marginal performance gains, it incurs prohibitive computational costs and unnecessary environmental impact. Furthermore, aggressive tuning falls outside the scope of this work: the primary objective of CiF is to establish global baseline trends and release a large-scale foundational dataset, rather than to overfit specific architectures to the domain.

Due to the scale of the dataset and the number of models evaluated, the experiments required substantial computational resources. All experiments were conducted on the IBM Research internal cluster. The main training phase for the 17 baseline models utilized 17 compute nodes, each equipped with 8 NVIDIA A100 GPUs, for approximately 1 day. Including preliminary experiments, and validation runs, the total project consumed approximately $10{,}000$ NVIDIA A100 GPU hours. All evaluation runs were performed on a single node with 8 NVIDIA A100 GPUs, requiring approximately 2 days.

\subsection{Class Co-occurrence}
\label{sec:appendix_cooccurrence}
\begin{figure}[h]
  \centering
  \includegraphics[width=0.65\linewidth]{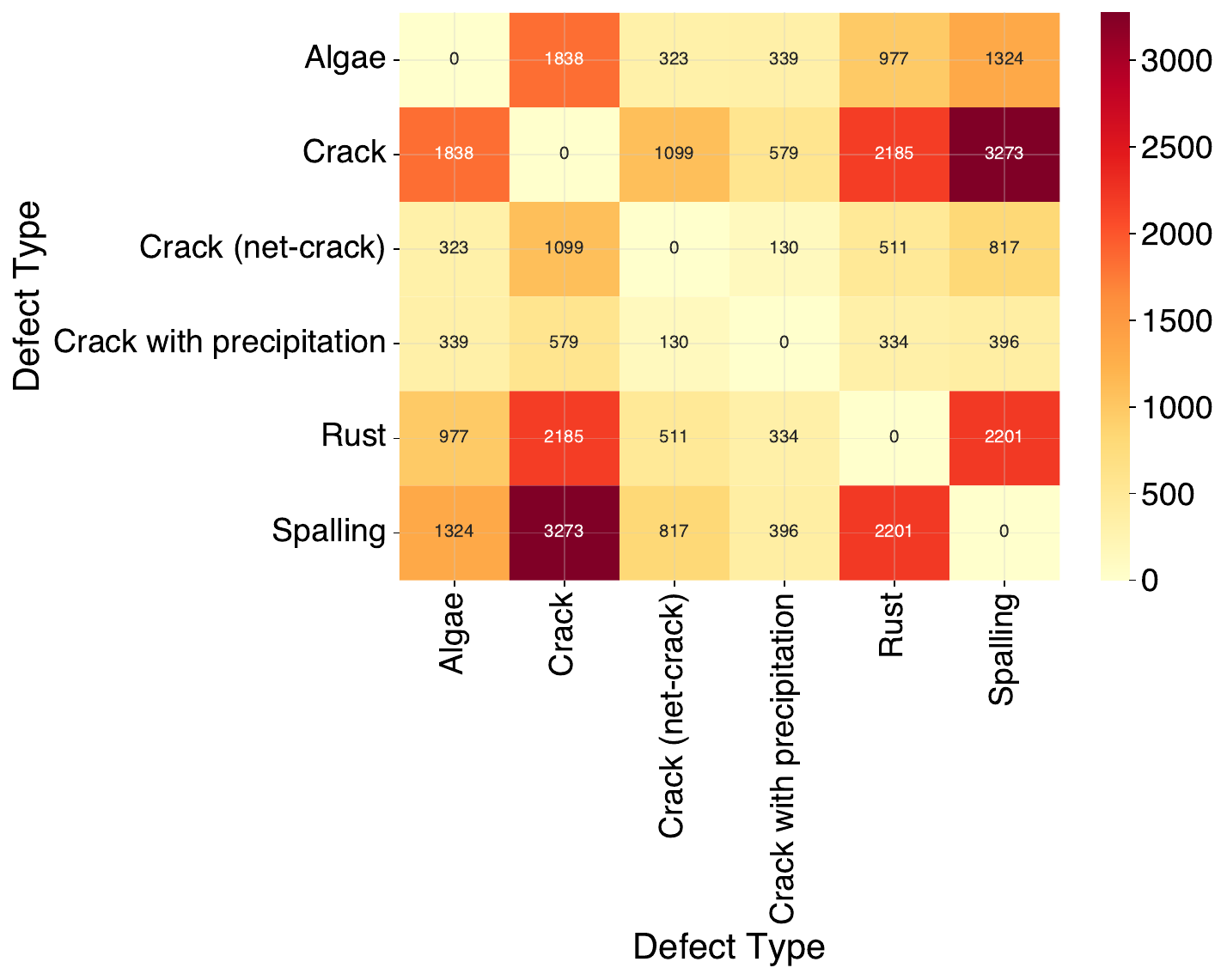}
  \caption{Co-occurrence frequency between defect categories within the same image (Full variant).}
  \label{fig:cooccurrence_full}
\end{figure}

\begin{table}[htb]
  \centering
  \small
  \setlength{\tabcolsep}{10pt}
  \caption{Per-class median compactness on the Full variant.}
  \label{tab:shape_metrics}
  \vspace{2pt}
  \begin{tabular}{lc}
    \toprule
    Class         & Compactness (median) \\
    \midrule
    Algae         & 0.69 \\
    Crack         & 0.58 \\
    Net-Crack     & 1.00 \\
    Precipitation & 0.29 \\
    Rust          & 0.68 \\
    Spalling      & 1.00 \\
    \bottomrule
  \end{tabular}
\end{table}
Figure~\ref{fig:cooccurrence_full} reports the frequency with which pairs of defect categories appear in the same image (Full variant). Crack and Spalling co-occur most frequently (3,273 images), followed by Rust and Spalling (2,201) and Crack and Rust (2,185), illustrating the presence of multiple defect types within individual scenes.

\subsection{Per-class Shape Compactness}
\label{sec:appendix_shape}
Compactness is the ratio of the instance area to the area of its convex hull, taking values in $[0, 1]$ with values close to $1$ indicating convex, blob-like shapes. Table~\ref{tab:shape_metrics} reports per-class median values on the Full variant. Net-Crack and Spalling are the most compact categories (median 1.00), consistent with their tendency to form simple, convex geometries or polygonal networks. In contrast, Crack with Precipitation has the most irregular boundaries (0.29), driven by diffuse deposition patterns. Crack sits slightly above the midpoint (0.58), reflecting the irregular, branching geometry of individual fractures.

\subsection{Full Resolution Experiments}
\label{sec:full_res}
Our images average 16.66 Megapixels and reach up to 42.18 Megapixels. We initially attempted to process them natively, but because a single full-resolution sample causes out-of-memory errors even on NVIDIA A100 80GB GPUs, we trained our "full resolution" baselines using the maximum downscaled input size that would fit into memory on a single A100. These effective dimensions are reported in the "Resolution" column of Table~\ref{tab:rebuttal-full-results}. We can see the general performance drop: as anticipated, the metrics are exceptionally poor across the entire evaluation when models are forced to handle these massive resolutions. Between the evaluated models, YOLO models perform slightly better in this regime, most likely because their architecture is inherently more scalable to high-resolution inputs.

\begin{table*}[htb]
\caption{Results on the Full variant of CiF. The $AP_{50:95}$ metric denotes instance segmentation accuracy for the first three supervised rows and SAM3, and object detection accuracy for the remaining models. Input dimensions are listed in the Resolution column.}
\label{tab:rebuttal-full-results}
\centering
\setlength{\tabcolsep}{1.2mm}
\resizebox{\linewidth}{!}{%
\begin{tabular}{l c c @{\hskip 5mm} c c c c c c}
\toprule
\multirow{2}{*}{Method} & \multirow{2}{*}{Resolution} & \multirow{2}{*}{$mAP_{50:95}$} & \multicolumn{6}{c}{$AP_{50:95}$} \\
\cmidrule(lr){4-9}
& & & Algae & Crack & Net-Crack & Precipitation & Rust & Spalling \\
\midrule
EoMT (Segmentation) & $1728 \times 2304$ & 5.6 & 15.4 & 0.5 & 5.0 & 0.9 & 7.6 & 4.6 \\
YOLOv11-m & $2388^2$ & 7.1 & 2.8 & 4.2 & 0.8 & 2.0 & 19.1 & 13.6 \\
YOLOv26-m & $2388^2$ & 10.6 & 15.2 & 6.3 & 3.4 & 2.3 & 20.8 & 15.3 \\
\midrule
RF-DETR (Detection) & $4192^2$ & 7.7 & 10.5 & 4.7 & 1.1 & 0.7 & 18.7 & 10.7 \\
YOLOv11-m & $2388^2$ & 10.9 & 3.6 & 12.1 & 1.3 & 5.0 & 24.1 & 19.2 \\
YOLOv26-m & $2388^2$ & 16.1 & 19.4 & 16.7 & 5.5 & 7.0 & 26.7 & 21.3 \\
\midrule
SAM3-Agent & $1024^2$ & 0.4 & 0.0 & 0.0 & 0.0 & 0.0 & 2.6 & 0.0 \\
SAM3-Text & $1024^2$ & 1.2 & 0.0 & 0.0 & 0.0 & 0.1 & 6.5 & 0.0 \\
SAM3-Oracle \#1 point & $1024^2$ & 3.8 & 6.3 & 0.0 & 0.5 & 0.7 & 9.9 & 5.5 \\
SAM3-Oracle \#3 points & $1024^2$ & 4.1 & 8.0 & 0.0 & 2.7 & 0.9 & 7.7 & 5.5 \\
SAM3-Oracle \#5 points & $1024^2$ & 4.3 & 8.8 & 0.0 & 3.3 & 2.2 & 6.9 & 4.8 \\
SAM3-Oracle Bounding Box & $1024^2$ & 21.2 & 34.3 & 0.2 & 31.2 & 7.0 & 29.4 & 24.9 \\
\midrule
Qwen3-VL-4B & full-resolution & 0.7 & 1.9 & 0.6 & 0.0 & 0.1 & 1.3 & 0.4 \\
Qwen3-VL-8B & full-resolution & 0.6 & 0.9 & 0.5 & 0.1 & 0.3 & 1.1 & 0.8 \\
Qwen3-VL-32B & full-resolution & 0.6 & 1.7 & 0.6 & 0.0 & 0.2 & 0.6 & 0.3 \\
\bottomrule
\end{tabular}
}
\end{table*}

\subsection{Annotation Quality Control}
\label{app:qa}

\textbf{Limitations of standard multi-annotation:} Standard dataset creation often relies on consensus across multiple independent annotators~\cite{gupta2019lvis}, but we found this conventional approach unworkable for infrastructure defects. Early in the project, we had multiple non-expert annotators independently label a subset of images, which resulted in notably low agreement (25\% recognition quality at 68\% segmentation quality). Crucially, this divergence was driven by disagreements over whether a defect was actually present, rather than how to draw its boundaries. Identifying structural anomalies requires specific domain expertise that a simple majority vote among standard annotators cannot replace. In light of this fundamental limitation, we designed a custom expert-review process for our needs, which sidesteps the pitfalls of independent consensus to reliably reduce residual errors. 

\textbf{Our QA procedure:}
An external vendor labeled each batch according to extensive annotation guidelines and conducted a first-pass quality assurance (QA) process. Our in-house team, trained by domain experts, then performed a second-pass review on a 10\% image sample per batch, scoring instance-level false positives (FP) and false negatives (FN). Batches above 10\% FP/FN were returned to the vendor for full re-annotation/correction and re-reviewed afterwards; batches below were accepted after correcting the found errors. Blurry, defocused, or badly exposed frames were discarded before annotation, so image blur is not a source of label error in the released dataset. Across all batches, our QA procedure produced 47{,}795 instance-level corrections over 2{,}922 re-annotated/corrected images.

\textbf{(1) Proportion of expert review:} In-house annotators trained by the domain experts reviewed a 10\% image sample per batch, which then served as the accept/reject trigger. Failing batches were fully re-annotated/corrected and re-reviewed, so the effective expert-touched fraction is substantially higher for the batches that needed it.

\textbf{(2) Accepted vs. rejected batches:} Of a total of 15 batches, eight were rejected on first review, with first pass error rates of up to 49\% FP and 42\% FN. They were then re-annotated and re-reviewed before acceptance. The rest were accepted after small corrections. Every released batch met the $\leq 10\%$ criterion at acceptance. The high first-pass error is precisely what the review loop is designed to catch, and none of it is present in the released data.

\textbf{(3) Realized error at acceptance:}
Accepted batches have a final error rate of about 3\% FP and 4\% FN on average, well below our strict 10\% threshold. We calculate these final numbers based on the last audit step right before a batch is approved. During this stage, our experts only correct the mistakes they find inside the 10\% audit sample. To estimate the remaining error, we project the mistake rate found in this sample onto the unreviewed 90\% of the batch. This projection assumes the 10\% sample accurately represents the whole batch. Naturally, these rates vary depending on the batch and the specific defect. Cleanly annotated batches with well-defined categories have error rates under 1\%. In contrast, the most visually ambiguous cases, such as easily missed rust or algae, can reach up to a 10\% FN rate.

\textbf{(4) Per-class boundary error.} To characterize boundary quality per class, we compare each released label against its pre-correction version on the re-annotated batches, using mask IoU and Boundary IoU (overlap within a 3 px band around the contour, which penalizes edge shifts that mask IoU hides). The results are presented in Table~\ref{tab:rebuttal-boundary-quality} where agreement is high for compact, well-defined classes (crack, spalling, rust; Boundary IoU $\geq 0.97$) and lowest for the wide-area or visually ambiguous classes (net-crack 0.79, algae 0.91), identifying these as the hardest to delineate.

\begin{table}[htb]
\centering
\small
\setlength{\tabcolsep}{10pt}
\caption{Per-class mask IoU and Boundary IoU between released labels and their pre-correction versions on re-annotated batches.}
\label{tab:rebuttal-boundary-quality}
\vspace{2pt}
\begin{tabular}{lcc}
\toprule
Class & Mask IoU & Boundary IoU (3px) \\
\midrule
Crack & 0.975 & 0.974 \\
Crack with Precipitation & 0.972 & 0.971 \\
Spalling & 0.988 & 0.976 \\
Rust & 0.991 & 0.978 \\
Algae & 0.969 & 0.908 \\
Net-Crack & 0.918 & 0.786 \\
\bottomrule
\end{tabular}
\end{table}

\textbf{(5) Ambiguous samples.} Class relabeling during review was rare: nearly all corrections were boundary refinements or missed instances rather than reassignments between classes. Ambiguity therefore surfaces as boundary disagreement, concentrated in the classes identified above. The residual ambiguous cases are presence decisions rather than extent decisions, principally whether a dark linear feature is a crack or a shadow.

\subsection{Zero-Shot and Foundation Model Evaluation Details}
\label{sec:appendix_prompt}

To evaluate out-of-the-box generalization rather than prompt-engineering sensitivity, all foundation models were tested using their official, minimally modified prompting protocols. 

\textbf{Qwen3-VL.} Inference was conducted via vLLM~\cite{kwon2023efficient} using greedy decoding to ensure deterministic outputs. Following the official Qwen3-VL 2D grounding cookbook, we queried the models with the standard prompt: \textit{``Locate every instance that belongs to the following categories: `algae, crack, net-crack, crack with precipitation, rust, spalling'. Report bbox coordinates in JSON format.''} Outputs were processed using the official Qwen3-VL parsing script. Formatting and parsing errors occurred in a negligible $0.3\%$ of test samples, confirming that the low detection performance reflects fundamental visual grounding limitations rather than output-parsing failures.

\textbf{SAM3.} For SAM3-Text, the model was prompted directly with the target class name. For SAM3-Agent, we retained the strictly unmodified system prompts governing its tool-use protocol (e.g., \texttt{segment\_phrase}, \texttt{examine\_each\_mask}) and iterative mask-verification sub-loop. Domain adaptation was restricted entirely to the user query: \textit{``the model should produce a mask for each \{category\} in this image of a civil infrastructure defect survey.''}


\end{document}